\documentclass[preprint,12pt,authoryear]{elsarticle}

\usepackage{amssymb}

\journal{Artificial Intelligence}

\usepackage{url}

\usepackage[ruled,vlined]{algorithm2e}

\usepackage[smaller]{acronym}
\usepackage{xspace}
\usepackage[table]{xcolor}
\usepackage{amsmath}
\usepackage{amsthm}
\usepackage{amssymb}
\usepackage{subcaption}

\usepackage{footnote}
\makesavenoteenv{table}

\usepackage{array}
\usepackage{multirow}
\usepackage{booktabs}

\usepackage{tikz}
\tikzset{
    agent/.style={
        thick, fill=blue!30, draw,
        rectangle, minimum width=0.4cm, minimum height=0.4cm
    },
    agent-head/.style={
        thick, fill=blue!50, draw,
        rectangle, minimum width=0.3cm, minimum height=0.15cm
    }
}

\usepackage{footmisc}

\newtheorem{example}{Example}

\acrodef{SAM}{Safe Action Model Learning}
\acrodef{PAC}{Probably Approximately Correct}
\acrodef{NSAM}{Numeric Safe Action Model Learning}
\newtheorem{definition}{Definition}

\newcommand{\nsam}{NSAM\xspace}

\newcommand{\mnsam}{$M_{\textit{NSAM}}$\xspace}

\newcommand{\sam}{\ac{SAM}\xspace}
\newcommand{\tuple}[1]{\ensuremath{\left \langle #1 \right \rangle }}
\newcommand{\pre}{\textit{pre}}
\newcommand{\eff}{\textit{eff}}
\newcommand{\name}{\textit{name}}
\acrodef{RL}{Reinforcement Learning}

\newcommand{\rl}{\ac{RL}\xspace}
\acrodef{IL}{Imitation Learning}
\newcommand{\il}{\ac{IL}\xspace}

\acrodef{BC}{Behavioral Cloning}
\newcommand{\bc}{\ac{BC}\xspace}

\acrodef{GAIL}{Generative Adversarial Imitation Learning}
\newcommand{\gail}{\ac{GAIL}\xspace}

\acrodef{PPO}{Proximal Policy Optimization}
\newcommand{\ppo}{\ac{PPO}\xspace}

\acrodef{RAMP}{Reinforcement learning, Action Model learning, and Planning}
\newcommand{\hybrid}{\ac{RAMP}\xspace}

\acrodef{DQN}{Deep Q-Network}
\newcommand{\dqn}{\ac{DQN}\xspace}
\newcommand{\qrdqn}{\ac{QR-DQN}\xspace}
\acrodef{QR-DQN}{Quantile Regression DQN}

\newcommand{\mcraft}{Minecraft\xspace}
\newcommand{\pogotask}{Craft Wooden Pogo\xspace}
\newcommand{\swordtask}{Craft Wooden Sword\xspace}
\newcommand{\bothtasks}{\swordtask and \pogotask\xspace}
\newcommand{\full}{{All Blocks}\xspace}

\newcommand{\pddla}{{$\text{NSAM}_{(+p)}$}\xspace}
\newcommand{\pddlaT}{{$\text{NSAM}_{(+pt)}$}\xspace}
\newcommand{\actions}{\textit{actions}}
\newcommand{\sfrom}{\ensuremath{s_{\textit{from}}}\xspace}
\newcommand{\sto}{\ensuremath{s_{\textit{to}}}\xspace}

\newcommand{\hybrido}{{\ac{RAMP}$_{(-p)}$}\xspace}
\newcommand{\hybridl}{{\ac{RAMP}$_{(-pn)}$}\xspace}

\begin{document}
\begin{frontmatter}

\title{Integrating Reinforcement Learning, Action Model Learning, and Numeric Planning for Tackling Complex Tasks}

\author[label1]{Yarin Benyamin}
\ead{bnyamin@post.bgu.ac.il}
\author[label1]{Argaman Mordoch}
\author[label1]{Shahaf S. Shperberg}
\author[label1]{Roni Stern}
\affiliation[label1]{organization={Software and Information Systems Engineering, Ben-Gurion University of the Negev},city={Be'er Sheva},
country={Israel}}

\begin{abstract}

Automated Planning algorithms require a model of the domain that specifies the preconditions and effects of each action. Obtaining such a domain model is notoriously hard. 
Algorithms for learning domain models exist, yet it remains unclear whether learning a domain model and planning is an effective approach for numeric planning environments, i.e., where states include discrete and numeric state variables. 
In this work, we explore the benefits of learning a numeric domain model and compare it with alternative model-free solutions. 
As a case study, we use two tasks in \mcraft, a popular sandbox game that has been used as an AI challenge. 
First, we consider an offline learning setting, where a set of expert trajectories are available to learn from. 
This is the standard setting for learning domain models. 
We used the 
Numeric Safe Action Model Learning (\nsam) algorithm to learn a numeric domain model and solve new problems with the learned domain model and a numeric planner. 
We call this model-based solution \pddla, and compare it to several model-free \il and Offline \rl algorithms. Empirical results show that some \il algorithms can learn faster to solve simple tasks, while \pddla allows solving tasks that require long-term planning and enables generalizing to solve problems in larger environments. 
Then, we consider an online learning setting, where learning is done by moving an agent in the environment. 
For this setting, 
we introduce the \hybrid hybrid strategy. 
In \hybrid, observations collected during the agent's execution are used to simultaneously train an \rl\ policy and learn a planning domain action model. 
This forms a positive feedback loop between the \rl policy and the learned domain model. 
We demonstrate experimentally the benefits of using \hybrid, showing that it finds more efficient plans and solves more problems than several \rl baselines. 

\end{abstract}

\end{frontmatter}

\section{Introduction}

\acresetall

Automated Planning is a state-of-the-art approach for solving goal-oriented sequential decision-making problems. Automated Planning algorithms do not require interactions with the environment to decide which actions to perform to achieve the desired goals. 
Instead, they rely on a \emph{domain model} that defines the preconditions and effects of the agent's actions. 
Using a domain model allows some planning algorithms to have formal guarantees on their behavior and output, which provides reliability and predictability. 
The dependency on a domain model is also a significant limitation since obtaining such a model can be very difficult, even for human experts.

To mitigate this limitation, prior work proposed automated methods for learning domain models~\citep{juba2021safe,wang1994learning,aineto2019learning,Lamanna24,XiGT24,MordochSSJ24} from observations. 
While these learning methods have shown some promise, their application beyond standard automated planning benchmarks remains limited~\citep{chitnis2021glib, jin2022creativity, james2022autonomous, sreedharan2023optimistic}.
This is especially true in \emph{numeric planning environments}, which are environments that include both discrete and numeric state variables. 
Indeed, there are very few algorithms for learning numeric planning action models~\citep{mordoch2023learning,segura2021discovering}, and numeric planners are less developed than planners for classical, discrete, environments. 
Therefore, one may consider using an alternative model-free approach to solve numeric planning problems. 
Depending on the exact problem setting, one may use \il and \rl algorithms as alternatives. 
These types of algorithms follow either a model-free or model-based approach. In model-free methods, observations are used to learn a state-action value function, which helps derive a policy, or to learn the policy directly—without attempting to model the environment’s dynamics. In contrast, model-based approaches aim to learn both the transition dynamics and the policy. These methods typically rely on function approximators, usually neural networks, to model the dynamics as a probability function. However, they do not leverage expressive and structured representations like PDDL, which can provide a more interpretable and generalizable understanding of the environment.
\textbf{In this work, we explore the benefits of learning a domain model for solving numeric planning problems, and compare it with alternative model-free approaches.}

As a case study, we use two tasks in the Polycraft World AI Lab (PAL) environment~\citep{palucka_2017,goss2023polycraft}. 
PAL provides a symbolic wrapper to \mcraft, a widely popular video game. 
We intentionally do not use domains from the International Planning Competition (IPC)~\cite{taitler2024ipc}. 
These domains are common benchmarks for automated planning algorithms, but have rarely been used to evaluate model-free algorithms. 
In contrast, the \mcraft environment has been used by prior work on model-based and model-free algorithms~\citep{roberts2017automated,benyamin2024crafting,tessler2017deep,lifshitz2024steve,dreamerv3}.
\textbf{The first contribution of this work is a full definition of two benchmark \mcraft tasks, along with source code and problem generators that allow running standard planning and \rl algorithms.}\footnote{We have described parts of this benchmark in our preliminary work~\cite{benyamin2024crafting}. However, in this work, we describe it in greater detail and include an additional task.}

The first problem setting we consider in this work is \emph{offline learning}, where a set of expert trajectories are available to learn from. 
These trajectories are created by executing a plan in the environment that successfully achieves an intended goal. 
The objective is to use these trajectories to learn how to generate plans for new problems in the same domain. 
This offline learning setting is the standard setting for domain model learning algorithms. 
The expert trajectories are given to a domain model learning algorithm, which outputs a domain model that is then used by an off-the-shelf numeric planner to solve new problems in the same environment. 
In our experiments we used 
\nsam~\citep{mordoch2023learning} to learn the numeric planning domain and MetricFF~\cite{hoffmann2003metric} to plan with the learned model. 
We refer to this model-based solution as \pddla. 

\textbf{The second contribution of this work is a comparison of this model-based solution and model-free alternative.} 
Specifically, we compared \pddla with offline \rl algorithms, namely \dqn~\citep{mnih2013playing} and \qrdqn~\cite{DabneyRBM18}, 
and two Imitation Learning algorithms, namely \bc~\citep{pomerleau1988alvinn} and \gail~\citep{ho2016generative}. 
A somewhat similar comparison of planning with a learned domain model and model-free alternatives was done in the context of 
classical planning environments~\citep{sreedharan2023optimistic}, but never before for numeric planning environments. 

Our results show that most \il and offline \rl algorithms were not competitive with \pddla, as they struggled to learn effectively from a limited number of trajectories.
Nevertheless, one \il algorithm, namely \bc, was able to learn to solve simpler problems faster than \pddla. For more complex problems that required longer-term planning, however, \pddla was able to learn much faster to solve problems than all other model-free alternatives. In addition, \pddla allowed seamless generalization to solve problems in larger environments.

The second problem setting we consider in this work is \emph{online learning}, where trajectories are collected by moving an agent in the environment. While there are algorithms for online learning of classical planning domains~\cite{lamanna2021online}, there is no such algorithm for online learning of numeric planning domains. 
\textbf{The third contribution of this work is a novel \hybrid  hybrid strategy.} 
\hybrid leverages \rl as a methodological means to explore the environment and gather observations, achieving a balance between exploration and goal-oriented exploitation. 
The observations collected during this process are used to simultaneously train a \rl policy and learn a numeric planning domain model. 
\hybrid uses the learned  domain model to attempt to find plans by calling an off-the-shelf domain-independent numeric planner. 
If a plan cannot be found, \hybrid acts according to the \rl policy and uses the learned domain model to try to find shortcuts in the plans it finds. 
A key feature of \hybrid is that the plans found by using the domain model are given to the \rl training process, which speeds up its learning process and yields higher-quality policies. 
 
We evaluated \hybrid in our \mcraft benchmark tasks and compared it to several standard \rl algorithms, including variants of \ppo~\citep{abbeel2004apprenticeship} and \dqn~\citep{mnih2013playing}. 
We observed that \hybrid significantly outperforms the best \rl algorithm we considered, solving more problems and finding, in some cases, plans that are 2 orders of magnitude shorter than the compared \rl algorithms. 

\section{Background}
\label{sec:background}
In this section, we provide relevant background in numeric planning, learning domain models for planning, and \rl. 
We also provide background on \mcraft, which is the environment we used in our experimental results. 

\subsection{Numeric Planning}
Numeric planning generally refers to solving planning problems in domains where 
action outcomes are deterministic, 
states are fully observable,
and the states are described with discrete and continuous variables. 
Numeric planning problems can be defined using extensions of the Planning Domain Definition Language (PDDL)~\citep{aeronautiques1998pddl}, such as  PDDL2.1~\citep{fox2003pddl2}. 
A numeric planning domain can be defined by a tuple $D=\tuple{F,X,A}$ 
where $F$ is a finite set of Boolean variables, 
$X$ is a set of numeric variables, 
and $A$ is a set of actions. 
A state is an assignment of values to all variables in $F\cup X$. 
Every action $a\in A$ is defined by a tuple $\tuple{\name(a),\pre(a),\eff(a)}$
representing the action's name, preconditions, and effects, respectively.  
Preconditions are assignments over the Boolean variables and conditions over the numeric variables, specifying when the action can be applied.  
The effects of an action are a set of assignments over $F$ and $X$, representing how the state changes after applying $a$. 
The set of actions with their definitions is referred to as the \emph{action model} of the domain. 
For a state $s$, action $a$, and action model $M$, we denote by $a_M(s)$ the state resulting from applying $a$ in state $s$ according to the effects specified by $M$. We omit $M$ in this notation when it is clear from the context.  
A planning problem in PDDL is defined by $\tuple{D,s_0, G}$ where $D$ is a domain, $s_0$ is the initial state, and $G$ are the problem goals. The problem goals $G$ are assignments of values to a subset of the Boolean variables and a set of conditions over the numeric variables. A goal state is a state that satisfies $G$. 
A solution to a planning problem is a \emph{plan}, i.e., a sequence of actions applicable in $s_0$ and resulting in a goal state.
Planning problems and domains are often specified in a \emph{lifted} form. 
This means that a planning domain defines parameterized predicates and functions, rather than Boolean and numeric variables, respectively, and the actions are parameterized actions as well. 
A planning problem defines a set of objects that can be used as parameters for these predicates, functions, and actions. 
For example, in the \mcraft\ domain we consider in our experiments, the location of the agent is a parameterized predicate $(at ?cell)$, and the problem specifies an object for every cell the agent can occupy. 


Numeric planning is, in general, a very hard problem computationally. 
From the computational complexity perspective, numeric planning is undecidable~\citep{helmert2002decidability}, 
even when restricted to having numeric conditions that involve one numeric
variable per formula and numeric effects that allow only the
addition of constants~\citep{gnad2023planning}. 
Some restricted classes are decidable but are still intractable~\citep{shleyfman2023structurally, bonet2020qualitative, gigante2023compilability}.

Despite these negative computational complexities, several domain-independent numeric planning algorithms have been proposed and deployed in practical applications over the years. 
Metric-FF~\citep{hoffmann2003metric}, ENHSP~\citep{scala2016interval}, Numeric Fast Downward~\citep{aldinger2017interval} and Nyx~\citep{piotrowski2024real}, 
are examples of such numeric planners. 
These planners usually work by employing a best-first search with different heuristic functions.

\subsection{Learning Domain Models for Planning}

Automated planning algorithms assume the existence of a symbolic model of the actions' preconditions and effects. 
However, obtaining a symbolic model is challenging, leading to the development of automated action model learning algorithms that learn from observing the agent's behavior.
In this work, we explore two settings of learning, offline and online.
\emph{Offline} domain model learning algorithms accept observations and output a domain model.
 \emph{Online} domain model learning algorithms choose which actions to perform in order to collect new observations and use these observations to continuously update and refine the domain model they eventually return.
  
 Algorithms for learning domain models vary in the assumptions they make on the available observations and the guarantees they provide on the action model they return~\citep{cresswell2013acquiring,amir2008learning,yang2007learning,aineto2019learning,juba2021safe}. 
 In Section~\ref{sec:related-work} we provide a brief overview of existing action model learning algorithms and a summary of their properties and assumptions.  

 To the best of our knowledge, there is no online action model learning algorithm that returns a numeric action model. The only offline action model learning algorithms capable of learning a numeric action model are the Numeric Safe Action Model Learning (\nsam) algorithms~\citep{mordoch2023learning} and PlanMiner~\citep{segura2021discovering}.\footnote{Some other action model learning algorithms are capable of learning the numeric costs or rewards associated with actions~\citep{jin2022creativity}, but they cannot learn numeric precondition and more complex numeric effects.}  

PlanMiner is a domain model learning algorithm that can learn a numeric action model from noisy observations. 
It uses symbolic regression to fit arithmetic formulas that represent the effects of observed actions. Using these formulas, it enriches the available data it has and learns the final preconditions and effects by calling NSLV~\citep{gonzalez2009improving}, a genetic algorithm for rule extraction. 
Solving a symbolic regression problem is, in general, computationally intractable. 
Thus, the runtime of PlanMiner may be prohibitively large. In addition, PlanMiner does not provide any guarantee on the correctness of the domain model it returns. 
Therefore, we chose to use \nsam in this work. 

\nsam makes several simplifying assumptions that allow it to run in polynomial time and provide strong safety guarantees over the domain model it returns. 
Specifically, it assumes full, noise-free, observability, 
and that preconditions are conjunctions of linear inequalities while effects are conjunctions of linear equations involving numeric state variables. 
Under these assumptions, \nsam is guaranteed to return a \emph{safe domain model}.  
\begin{definition}[Safe Domain Model]
A domain model is called safe if 
every plan consistent with it is  \emph{sound} with respect to the real, unknown action model.
\label{def:safe}
\end{definition}
\noindent Definition~\ref{def:safe} is a direct corollary from the exact definition of a safe action model given by~\cite{mordoch2023learning} and ~\cite{juba2021safe}.

Since \nsam plays a crucial role in this work, we elaborate below on how it works for completeness. 
\nsam decomposes the learning task into two steps: (1) learning the discrete part of the action model, and (2) learning the continuous part of the action model. 
To learn the discrete part of the action model, \nsam\ uses the \sam algorithm~\citep{juba2021safe}, which is based on applying a simple set of inductive rules on the given observations to identify which predicates are part of the actions' preconditions and effects. 
The algorithm initially assumes for every action that every relevant predicate is part of its preconditions and that it has no effects. 
Then, it processes the available trajectories to remove predicates that cannot be preconditions and adds predicates that must be effects of observed actions. 
\nsam learns numeric preconditions by determining the convex hull of observed numeric variables. 
It learns numeric effects by observing state variable changes before and after executing an action, formulating these changes as a system of linear equations, and finding a solution for the system.

\subsection{Reinforcement Learning and Imitation Learning}
\rl is a field of AI in which agents learn to make decisions by interacting with an environment and receiving feedback in the form of rewards (or penalties).
There are two types of \rl algorithms: \emph{off policy} and \emph{on policy}. 
Off-policy algorithms are algorithms that can learn from data collected by any policy and on-policy algorithms learn only from data collected by the current policy.
\dqn~\citep{mnih2015human} and \ppo~\citep{schulman2017proximal} are prominent examples of off-policy and on-policy \rl algorithms, respectively. 
Each algorithm has its pros and cons, but \ppo is renowned for its robust performance even in the absence of extensive hyperparameter tuning.

\il~\citep{schaal1999imitation} is a related research field where an AI agent is trained to ``mimic human behavior in a given task''~\citep{hussein2017imitation}. 
Two \il algorithms we discuss in this work are \bc\citep{pomerleau1988alvinn} and \gail~\citep{ho2016generative}. 
\bc employs supervised learning to mimic the expert's policy, while \gail takes a unique adversarial approach by simultaneously training a policy and a discriminator. The discriminator's role is to distinguish between expert observations and those generated by the learned policy.
Offline \rl~\citep{sutton2018reinforcement} is similar to \il except that the given trajectories encompass not only the states and actions but also the rewards associated with each transition. 
Offline \rl algorithms aim not to mimic these trajectories but to learn from them how to maximize future rewards. Consequently, off-policy algorithms, such as \dqn, can also serve as an offline \rl algorithm.

\section{Planning Benchmarks in Minecraft}
Automated planners are often evaluated on planning problems from the International Planning Competition (IPC)~\citep{taitler2024ipc}. 
Thus, they may be inadvertently optimized towards solving these types of problems. 
To provide a fairer comparison between model-based and model-free approaches, the experiments in this work were deliberately performed in an environment that is not part of the IPC, namely \mcraft. 
\mcraft is an open-world game in which we control a character named Steve that is situated within a 3D environment. Steve can do a variety of actions, including moving in the environment, mining resources, collecting items, crafting new types of items, building structures, and fighting enemies. 

Due to its widespread popularity and complex open-world setting, \mcraft has been used as a sandbox for developing and evaluating AI agents. 
Various wrappers have been created to enable AI agents to interact with the game. 
MineRL~\citep{guss2019minerl} is a \mcraft-based platform for the development, testing, and evaluation of \rl algorithms. 
It is compatible with OpenAI-Gym~\citep{brockman2016openai}, a common platform for \rl algorithms. 
MineRL is not suitable for our work as we do not consider a visual, pixel-based representation of the game. 

Instead, we used Polycraft~\citep{palucka_2017}, an interface to \mcraft that is part of the Polycraft World AI Lab (PAL)~\citep{goss2023polycraft}\footnote{https://github.com/PolycraftWorld/PAL}. 
PAL allows AI agents to easily interact with \mcraft's environment by sending commands to the API and waiting for a response. Each command has predefined preconditions, effects, and costs. This mechanism enables \rl algorithms to use the API to train their agents and attempt to solve various tasks. 
Unlike MineRL, PAL supports \emph{symbolic} observations, making it a better fit for our purposes. To facilitate comparison with \rl algorithms, we also developed an OpenAI-Gym wrapper for PAL, enabling seamless execution of any algorithm compatible with the OpenAI-Gym API.

\subsection{Planning Tasks}
For our evaluation, we used two tasks, namely \bothtasks, within the PAL \mcraft environment~\citep{goss2023polycraft}.
\begin{figure}[tbp]
    \centering
    \includegraphics[width=0.7\textwidth]{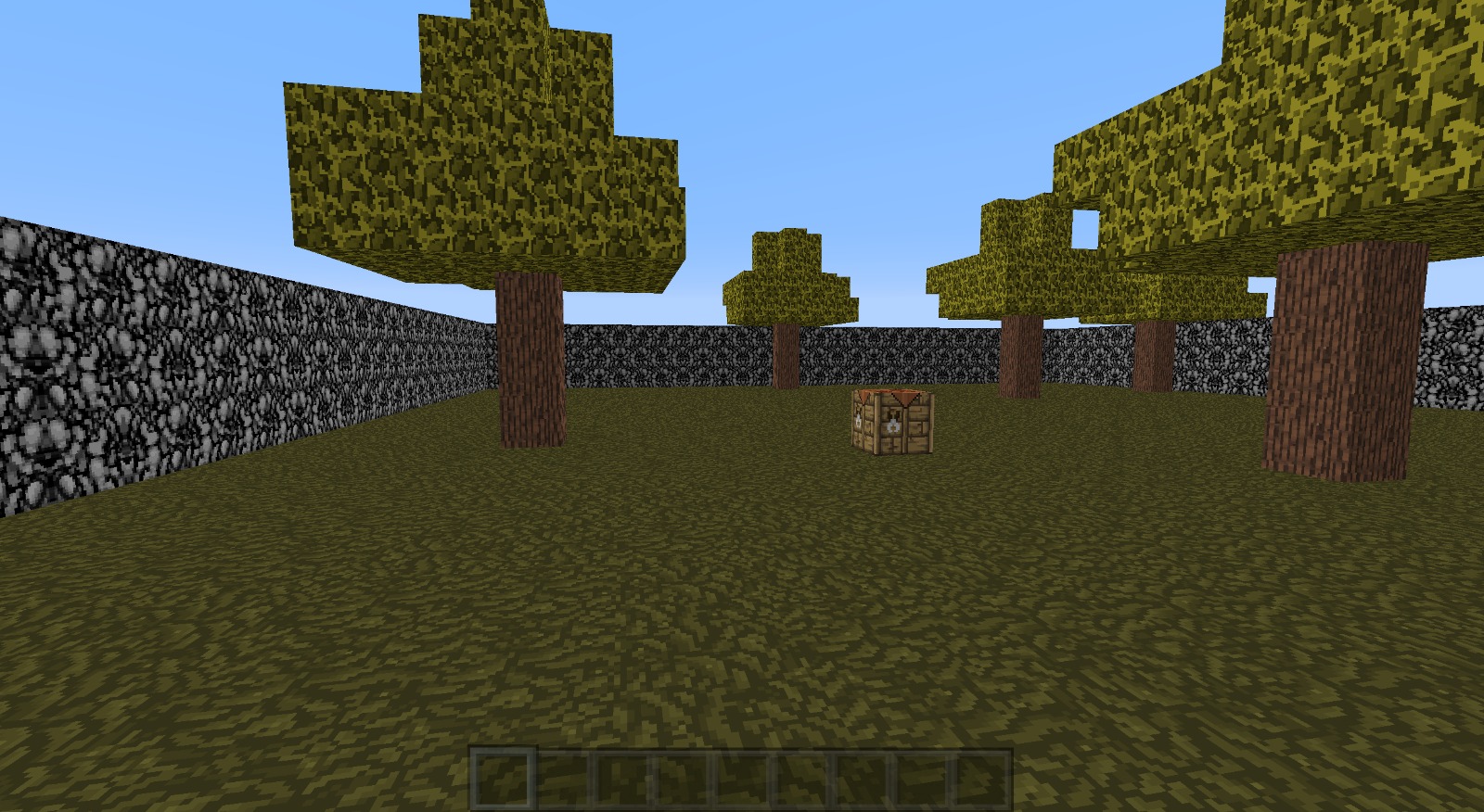}
    \caption{Example of Steve point-of-view at random map configuration}
    \label{fig:steve_pov}
\end{figure}
In both tasks, Steve is located in a field comprising $N \times N$ blocks and surrounded by unbreakable bedrock walls. 
The field includes multiple trees and a crafting table, as shown in Figure~\ref{fig:steve_pov}. 
The initial input includes the map data, i.e., what exists in each cell, as well as the number of items it is currently holding in the inventory. We consider two \mcraft tasks in this environment. 
\begin{figure}[tbp]
    \centering
    \includegraphics[width=0.7\columnwidth]{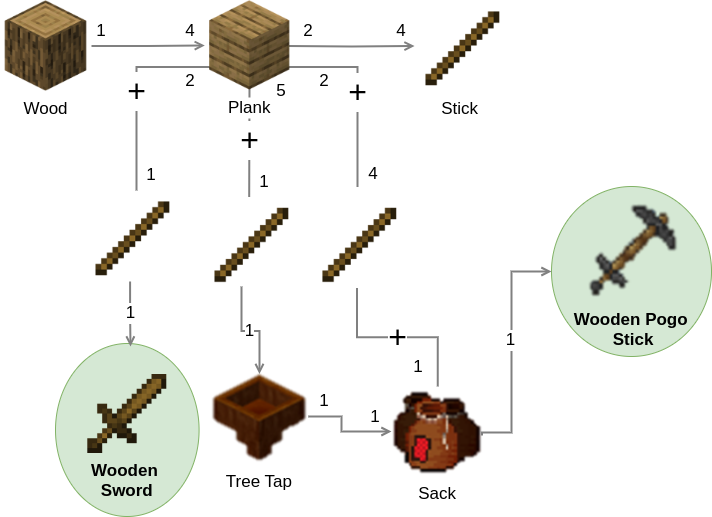}
    \caption{The crafting recipes required to accomplish the \bothtasks tasks.}
    \label{fig:tasks_route}
\end{figure}

\paragraph{Task 1: \swordtask} In this task, Steve must:
\begin{enumerate}
    \item Harvest at least one wood block from trees.
    \item Use one wood piece to craft 4 planks.
    \item Use 2 planks to craft 4 sticks.
    \item Use one stick and 2 planks to craft a wooden sword.
\end{enumerate}

\noindent The actions from the PAL API that are relevant here are:
\begin{enumerate}
\item \textbf{TP\_TO} - teleport from the current location to another cell on the map.
    \item \textbf{BREAK}- breaks a tree to extract and add the logs to the inventory.
    \item \textbf{CRAFT\_PLANK} - craft planks from the logs in the inventory.
    \item \textbf{CRAFT\_STICK} - craft sticks from the planks in the inventory.
    \item \textbf{CRAFT\_WOODEN\_SWORD} - teleport to the crafting table, craft a wooden sword, and add it to the inventory.
\end{enumerate}

\paragraph{Task 2: \pogotask} In this task, Steve must:
\begin{enumerate}
    \item Harvest at least three wood blocks from trees.
    \item Use 3 wood pieces to craft 12 planks.
    \item Use 4 planks to craft 8 sticks.
    \item Use one stick and 5 planks to craft a tree tap.
    \item Place the tree tap near a tree to collect polyisoprene sacks.
    \item Use 4 sticks, 2 planks, and one polyisoprene sack to craft a wooden pogo stick.    
\end{enumerate}

\noindent The actions from the PAL API that are relevant here are:
\begin{enumerate}
\item \textbf{TP\_TO} - teleport from the current location to another cell on the map.
    \item \textbf{BREAK}- breaks a tree to extract and add the logs to the inventory.
    \item \textbf{CRAFT\_PLANK} - craft planks from the logs in the inventory.
    \item \textbf{CRAFT\_STICK} - craft sticks from the planks in the inventory.
    \item \textbf{CRAFT\_TREE\_TAP} - teleport to the crafting table, craft one tree tap, and add it to the inventory.
    \item \textbf{PLACE\_TREE\_TAP} - when in front of a tree, move left, place the tree tap on it, collect a polyisoprene sack, and add it to the inventory.
    \item \textbf{CRAFT\_WOODEN\_POGO} - teleport to the crafting table, craft a wooden pogo stick, and add it to the inventory.
\end{enumerate}
Figure~\ref{fig:tasks_route} illustrates the recipes required to achieve both tasks. 
Observe that crafting actions are irreversible and the amount of resources in a map is fixed. Thus, the agent may end in a dead-end. For example, if all planks are crafted into sticks, the recipe for crafting a wooden pogo can no longer be completed. 
This emphasizes the importance of long-term planning in this domain. 

Note that we use ``teleport'' actions (TP\_TO) instead of moving the agent to the four cardinal directions as is done by a human player. 
These teleport actions are fully supported by the PAL framework~\cite{goss2023polycraft} and allow us to avoid the need for lower-level path planning. 
This is reminiscent of OpenAI's use of macro actions in Starcraft~\citep{vinyals2019grandmaster}.

\subsection{Problem Generators}
The original task in PAL was predefined with a fixed map size and predetermined positions for the trees, the crafting table, and the agent. In addition, Steve always started with an empty inventory. 
To generate more diverse initial states and problems, we developed a problem generator that randomizes: (1) the agent's starting position on the map, (2) the quantity and placement of trees, and (3) the items present in the agent's inventory. 
The problem generator also accepts as input the size of the desired map. 
Consequently, different sequences of actions are needed to solve different problem instances. 

In our experiments, we 
generated maps of sizes $6\times 6$, $10\times 10$, and $15\times 15$. 
The configured initial the number of items in the inventory to range from zero to eight for all items, except for the last major items required to achieve the goal item, which is a ``Stick'' for the \swordtask, and ``Tree Tap'' and ``Sack'' for the \pogotask.
These major items were always set to not be in the inventory of the initial state, to ensure that the planning problem is not trivial. 
We set the number of trees on each map to range from zero to (map size)/3.

\section{Offline Learning of Numeric Action Models}

In this setting, we are given an initial state $s$, a goal to achieve $g$, and a set of \emph{expert trajectories} $\mathcal{T}$ created by observing an \emph{expert} solve different problems in the same environment. The agent's objective in this setting is to learn from $\mathcal{T}$ how to output a \emph{plan} or a \emph{policy} for achieving $g$ from its current state ($s$). 
Learning and planning in this setting are done \emph{offline}, that is, before the agent performs any action in the environment. 
The main objective is to 
be able to solve other problems in the domain.
We consider two main approaches in this offline learning setting: a model-free approach and a model-based approach.

\subsection{Model-free approach} 
\il algorithms~\citep{schaal1999imitation} and Offline \rl algorithms~\citep{sutton2018reinforcement} are suitable for solving our offline learning setting. 
In our experiments, we implemented two \il algorithms, namely \bc~\citep{pomerleau1988alvinn} and \gail~\citep{ho2016generative}, and two Offline \rl algorithms, namely \dqn~\citep{mnih2013playing} and \qrdqn, an extension of \dqn which approximates the full distribution of returns using quantile regression, enabling better risk-sensitive decision-making and improved learning stability. The \il algorithms can be used as-is given the set of expert trajectories. 
To use an offline \rl algorithm, we must define a reward function that associates a state with the benefit of reaching it. 
In a goal-oriented task like ours, the natural way to define the reward function is for the agent to receive a positive reward if it achieves its goal, and zero otherwise. We define our reward function in this way: the agent receives a unit reward if it achieves the goal and zero otherwise.

To allow using off-the-shelf \rl and \il algorithms in our \mcraft benchmark, we implemented a translation mechanism that converts \mcraft states and actions into states and actions in a dedicated AI gym environment~\citep{brockman2016openai}. 
The resulting task is not easy since the number of actions in each state is $N^2 + 5$ for the \swordtask task and $N^2 + 6$ for the \pogotask task, where $N$ is the width and height of the map. 
Moreover, most actions are TP\_TO actions, which do not provide any positive reward.

\subsection{Model-based approach} 
In this approach, we learn a symbolic numeric planning domain model from the given expert trajectories $\mathcal{T}$ and use an automated planner to find a plan to achieve the goal. 
In our implementation, we provide the expert observations as input trajectories to \nsam~\citep{mordoch2023learning}, which outputs a PDDL domain model of the environment. 
Then, we used an off-the-shelf domain-independent numeric planner to solve the resulting planning problem and output a plan for the given task. 
In our experiments, we used Metric-FF~\citep{hoffmann2003metric}, a state-of-the-art numeric planner, but any domain-independent numeric planner can be used instead. 

\begin{figure}[ht]
    \centering
    \includegraphics[width=\columnwidth]{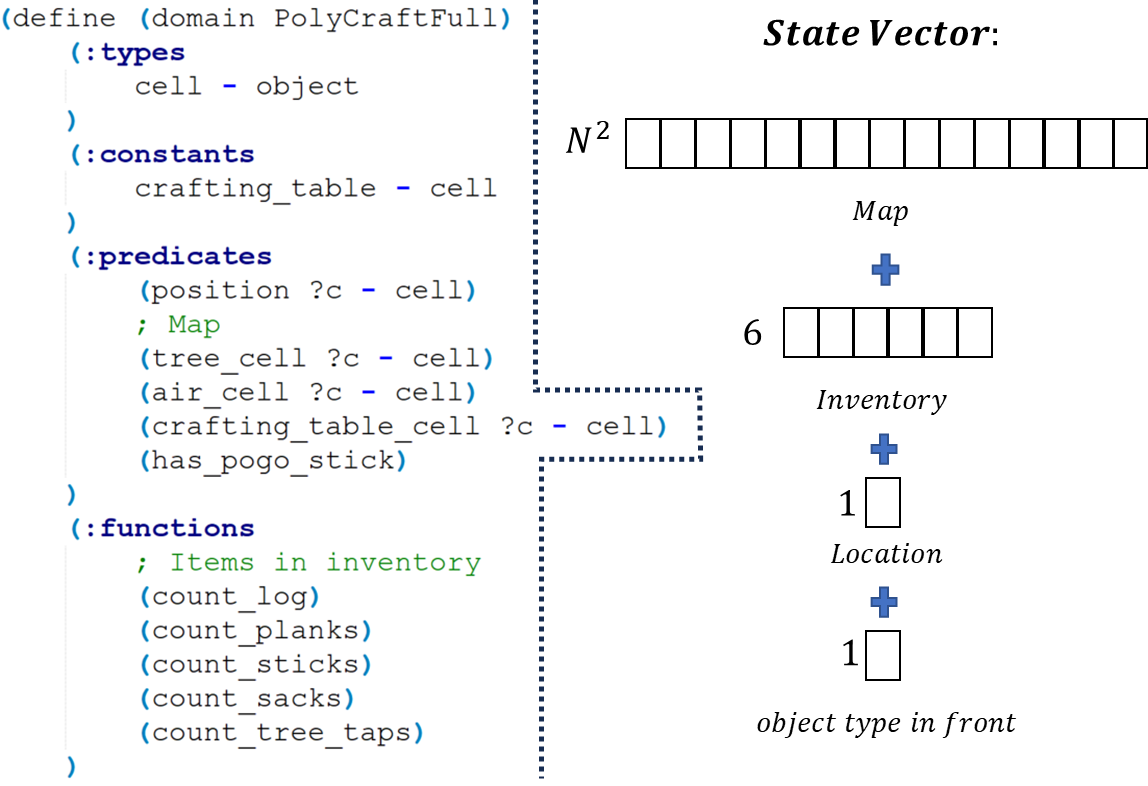}
    \caption{State representation of the \full model: PDDL (left), RL (right).}
    \label{fig:complex-actions-representation}
\end{figure}

\label{sec:translation-mechanism}
To use this approach in our \mcraft benchmark, we implemented a translation mechanism that converts \mcraft states and actions into corresponding PDDL2.1 states and actions (see Figure \ref{fig:complex-actions-representation}).
Specifically, each cell in the map is represented as an object, with the locations of Steve, the crafting table, and trees indicated by cell-based predicates, while inventory item counts are encoded as functions.

The resulting planning task is difficult because of the following reasons. 
First, the agent is not given the action model of the available actions, i.e., their preconditions and effects. 
Second, the action space of the underlying domain is very large. In particular, consider the TP\_TO action. In PDDL, this action involves two parameters: the current position of the agent and its target position. Consequently, grounding only this action results in $N^4$ grounded actions. 
Similarly, almost all other actions also require the agent's current position. Consequently, the total number of grounded actions in PDDL is $N^4 + 2N^2 + 2$ for the \swordtask task and $N^4 + 4N^2 + 2$ for the \pogotask task. This is a significantly larger action space than in our \rl models. Moreover, our gym environment includes an explicit specification of the cell type directly in front of Steve, which is a useful feature for some actions. Adding this feature as a state variable to our PDDL domain requires designing complex conditional effects.

\begin{figure}[ht]
    \centering        
    \includegraphics[width=0.9\linewidth]{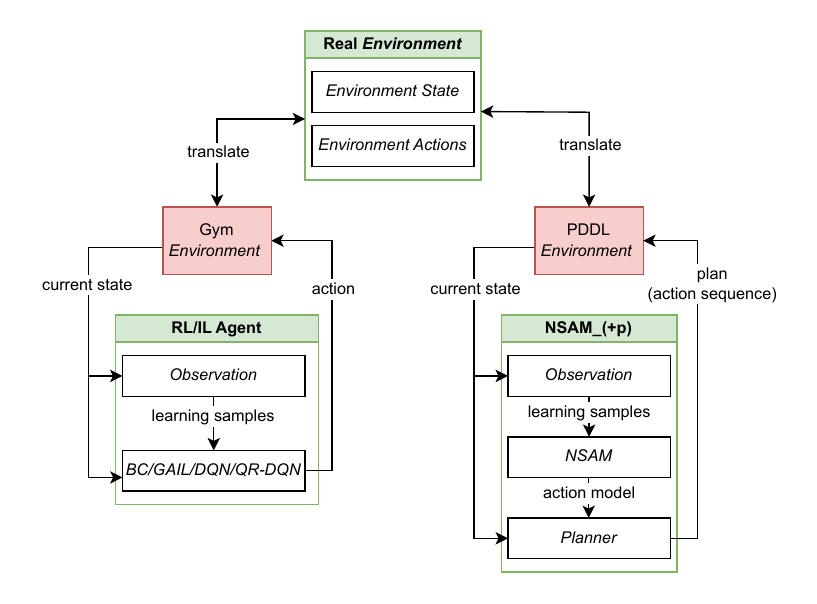}
    \caption{The evaluated approaches for the offline learning setting.}
    \label{fig:offline-approach}
\end{figure}

Figure~\ref{fig:offline-approach} summarizes \pddla and model-free approaches for solving our offline learning setting. 
The translation mechanisms described above enable a comparison of these approaches. 
They also allow using our experimental setup as a numeric planning benchmark, as well as an \rl benchmark, for future studies.

\subsection{Experimental Setup}
\label{sec:setup-offline}
We generated 1,000 instances of the 
\pogotask and 200 instances of the \swordtask for every map size ($6\times 6$, $10\times 10$, and $15\times 15$). 
We divided these instances into training and test sets with a 4:1 ratio. 
To ensure robustness and generalization, we repeated this process in a 5-fold split, calculating the average and standard deviation over the five folds.

For every instance in the training set, we generated a solution using an expert agent capable of solving the task. This expert agent was constructed by manually modeling the task as a numeric planning problem and employing MetricFF to solve it. Every generated plan was validated within the environment.

For every instance in the test set, we run \pddla, \bc, \gail, \dqn, and \qrdqn. 
If a goal was reached before performing 32 steps, we consider this run as successful. 
Otherwise, it is considered a failure. 
To ensure the experiment runtimes are manageable, we also imposed a time limit on the planner (for \pddla). 
If a plan was not found within that time limit, we consider the run as a failure. 
We set time limits of 5 seconds for all \swordtask problem instances, 30 seconds for the $6 \times 6$ and $10 \times 10$ \pogotask problem instances, and 3,600 seconds for the $15 \times 15$ \pogotask problem instances.
These time limits were chosen after a trial-and-error process, aiming for values that maintain the overall runtime of our experiments manageable while allowing the planner to find solutions for most cases. 
The main metric we consider is \emph{success rate}, which is the ratio of instances that were solved successfully. 

\paragraph{Implementation details} 
All the code used in our experiments is publicly available at \url{https://github.com/SPL-BGU/PolyPlan}. 
For our model-free algorithms, we used the implementation of \bc and \gail available in the \emph{imitation} 
library\footnote{\url{https://imitation.readthedocs.io}} and the implementation of \dqn and \qrdqn available in the \emph{stable\_baselines3} library
\footnote{\url{https://stable-baselines3.readthedocs.io}\label{ft:sb3}}. 
After some preliminary experiments and hyper-parameter tuning, we chose the following configuration. 
The neural network architecture consisted of 
three fully connected layers 
where the first layer has 512 units and the second and third layers have 256 units each. 
The hyperbolic tangent activation function was chosen for each layer.

\subsection{Experimental Results}
Despite exhaustive efforts in hyperparameter tuning and experimentation with various network architectures, \gail, \dqn, and \qrdqn consistently failed to solve the \bothtasks tasks across all experiments. 
Thus, we only show results for \pddla and \bc.

\begin{table}[ht]
\resizebox{\textwidth}{!}{
\centering
\begin{tabular}{|cc|ccccccc|ccc|}
\hline
\multicolumn{2}{|c|}{ } & \multicolumn{7}{c|}{\textbf{Wooden Pogo}} & \multicolumn{3}{c|}{\textbf{Wooden Sword}}\\
\hline
\textbf{} & \textbf{} & \multicolumn{7}{c|}{\textbf{Solution Length}} & \multicolumn{3}{c|}{\textbf{Solution Length}}\\
\textbf{Map}         & \textbf{ }       & \textbf{5} & \textbf{7} & \textbf{8} & \textbf{9} & \textbf{10} & \textbf{11} & \textbf{12+} & \textbf{2} & \textbf{3} & \textbf{5}\\
\hline
      & \# Prob.  & 123  & 216  & 145  & 164  & 146  & 86   & 120  & 109  & 78   & 13   \\
6x6   & BC     & \textbf{0.99} & 0.95 & \textbf{0.98} & 0.93 & 0.93 & 0.97 & 0.78 & \textbf{1.00} & 0.98 & 0.57 \\
      & \pddla  & 0.96 & \textbf{1.00} & \textbf{0.98} & \textbf{0.99} & \textbf{0.98} & \textbf{1.00} & \textbf{0.95} & 0.98 & \textbf{0.99} & \textbf{1.00} \\
\hline
      & \# Prob.  & 95   & 242  & 140  & 166  & 147  & 98   & 112  & 110  & 88   & 2    \\
10x10 & BC     & 0.77 & 0.78 & 0.65 & 0.60 & 0.44 & 0.46 & 0.34 & \textbf{1.00} & 0.94 & 0.00 \\
      & \pddla   & \textbf{0.97} & \textbf{0.97} & \textbf{0.97} & \textbf{0.99} & \textbf{0.99} & \textbf{1.00} & \textbf{0.96} & \textbf{1.00} & \textbf{0.98} & \textbf{1.00} \\
\hline
      & \# Prob.  & 108  & 243  & 143  & 164  & 137  & 99   & 106  & 126  & 66   & 8    \\
15x15 & BC     & 0.71 & 0.62 & 0.43 & 0.54 & 0.40 & 0.22 & 0.13 & \textbf{1.00} & 0.82 & 0.20 \\
      & \pddla   & \textbf{0.98} & \textbf{0.99} & \textbf{0.96} & \textbf{1.00} & \textbf{0.98} & \textbf{0.95} & \textbf{0.91} & \textbf{1.00} & \textbf{0.96} & \textbf{0.80} \\
\hline
\end{tabular}
}
\caption{Results for \pddla and \bc in the offline learning setting for \pogotask (left) and \swordtask (right).}
\label{tab:offline_results}
\end{table}

Table~\ref{tab:offline_results} shows the results after training with all the trajectories in the training set for all algorithms, map sizes, and tasks. 
To distinguish between easy and hard problem instances, we split the results (over the ``Solution Length'' columns) based on the solution length of the problem instances, as computed by the expert agent that generated the training set. 
The rows ``\# Problems'' show the number of problem instances for every solution length. 
The rows ``BC'' and ``\pddla'' show the success rate of each algorithm, averaged over the 5 folds.

The results show that \pddla is either as good or better than \bc in almost all cases. 
In fact, \bc is only better than \pddla in two places: problems with a solution length of 5 in \pogotask, where it has a success rate of 0.99 while \pddla only reaches 0.96, and problems with a solution length of 2 in \swordtask, where it has a success rate of 1.00 while \pddla only reaches 0.98. 
Note that these are also the easiest problems, i.e., those with the smallest solution length in each task. 
As expected, \pddla is significantly better than \bc in the harder problems, since it requires longer-horizon planning, which planners are designed for. 
For instance, for \swordtask problems with solution length 5, 
\bc solves only 57\% of test problems on the $6 \times 6$ maps and 20\% on the $15 \times 15$ maps, 
while \pddla solves all test problems on the $6 \times 6$ and 80\% of them on the $15 \times 15$ maps. 
The advantage of \pddla grows even more significantly in the \pogotask task, which is inherently more challenging than \swordtask.

\subsubsection{Zero-Shot Transfer Results}

\begin{table}[htbp]
    \centering
    \begin{tabular}{c c}
                \begin{minipage}{0.45\textwidth}
            \centering
            \begin{minipage}[b]{\textwidth}
                \includegraphics[width=\textwidth]{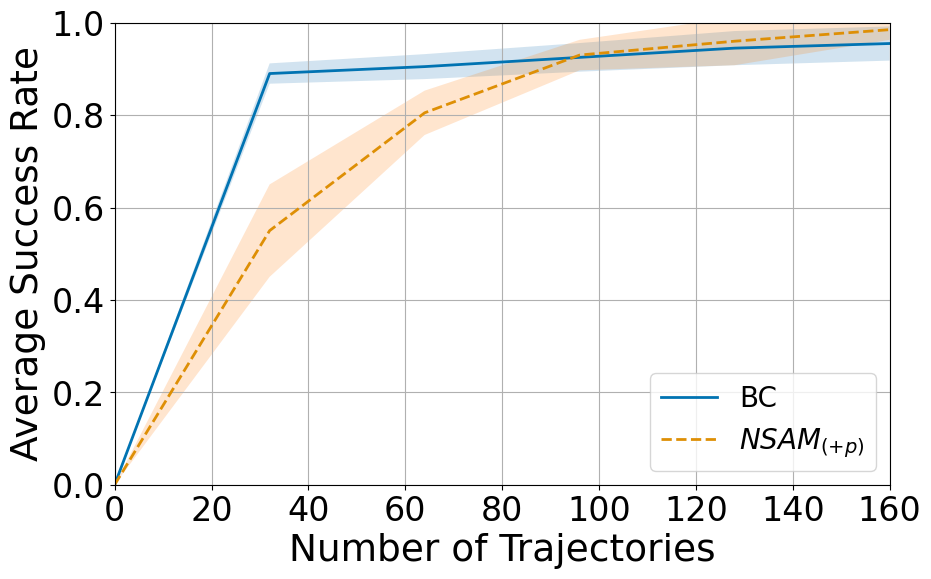}
                \captionof{figure}{\swordtask task on $6\times6$ maps}
                \label{fig:6x6-offline-sword}
            \end{minipage}
            \medskip
            \begin{minipage}[b]{\textwidth}
                \includegraphics[width=\textwidth]{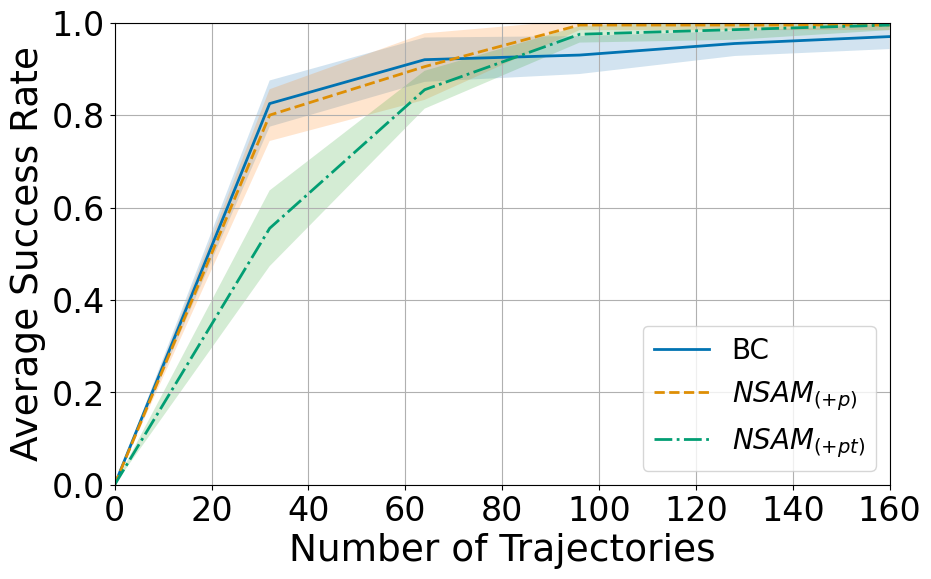}
                \captionof{figure}{\swordtask task on $10\times10$ maps}
                \label{fig:10x10-offline-sword}
            \end{minipage}
            \medskip
            \begin{minipage}[b]{\textwidth}
                \includegraphics[width=\textwidth]{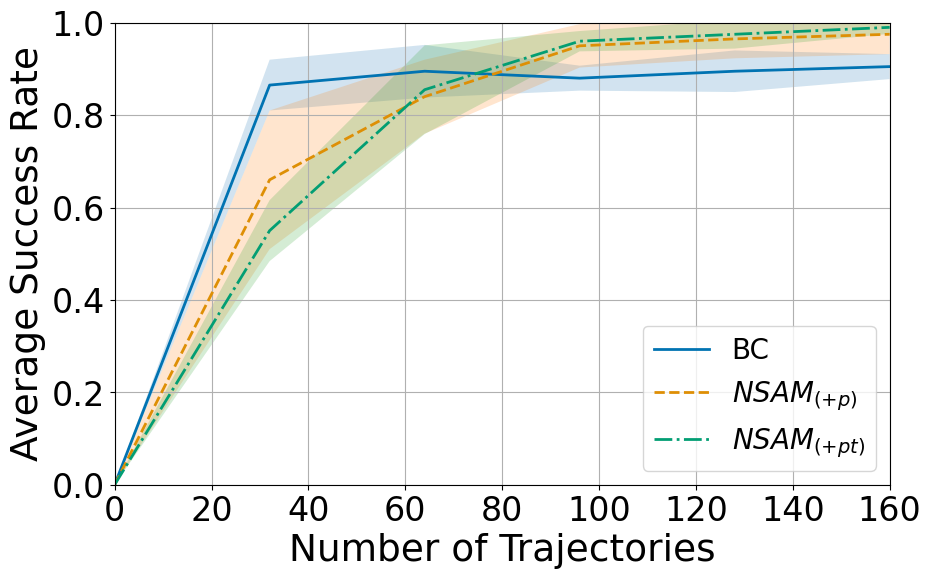}
                \captionof{figure}{\swordtask task on $15\times15$ maps}
                \label{fig:15x15-offline-sword}
            \end{minipage}
            \label{fig:offline_sword}
        \end{minipage}
        &
        \begin{minipage}{0.45\textwidth}
            \centering
            \begin{minipage}[b]{\textwidth}
                \includegraphics[width=\textwidth]{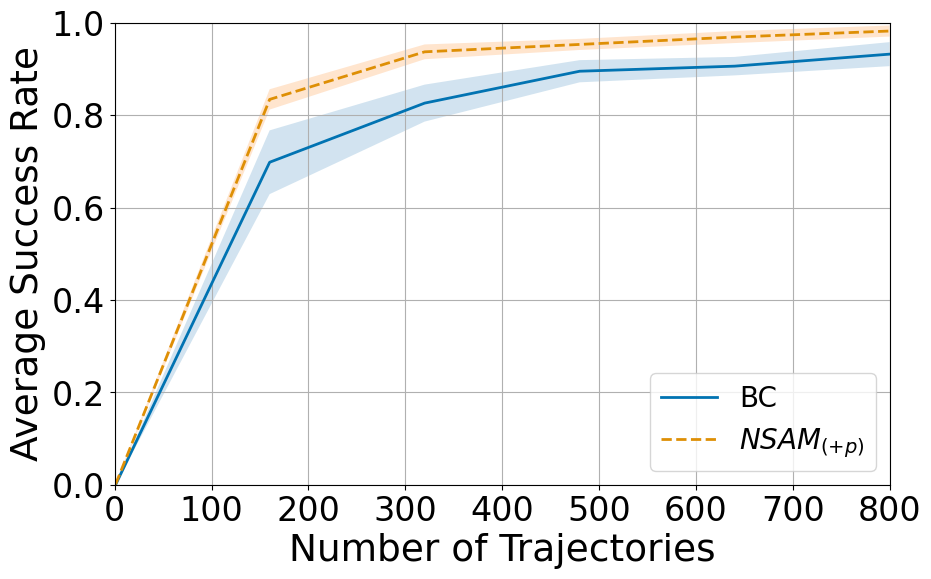}
                \captionof{figure}{\pogotask task on $6\times6$ maps}
                \label{fig:6x6-offline-pogo}
            \end{minipage}
            \medskip
            \begin{minipage}[b]{\textwidth}
                \includegraphics[width=\textwidth]{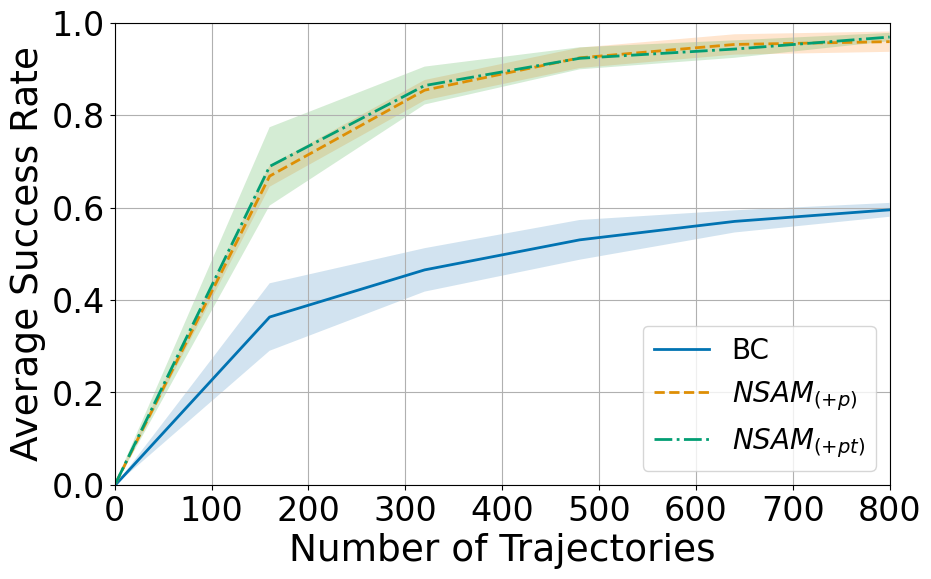}
                \captionof{figure}{\pogotask task on $10\times10$ maps}
                \label{fig:10x10-offline-pogo}
            \end{minipage}
            \medskip
            \begin{minipage}[b]{\textwidth}
                \includegraphics[width=\textwidth]{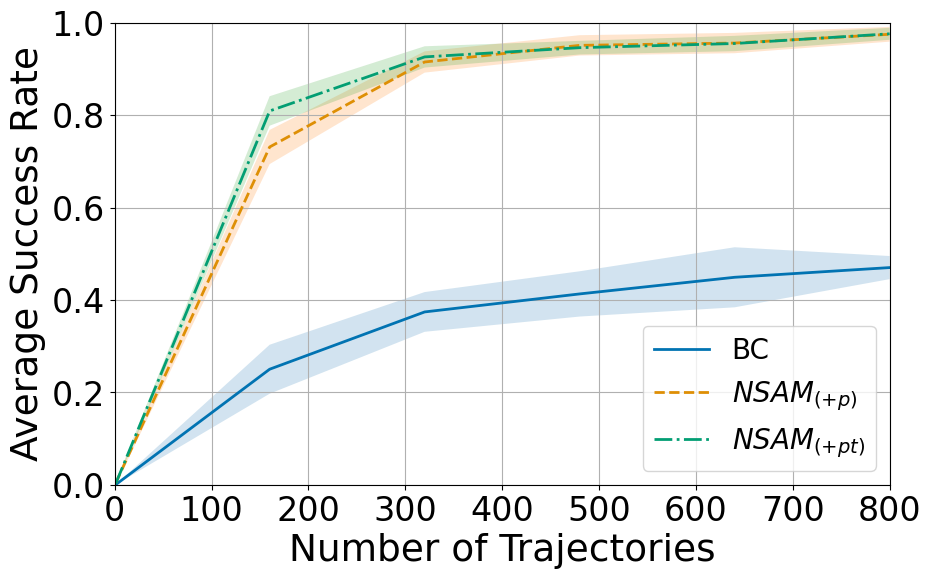}
                \captionof{figure}{\pogotask task on $15\times15$ maps}
                \label{fig:15x15-offline-pogo}
            \end{minipage}
            \label{fig:offline_pogo}
        \end{minipage}
    \end{tabular}
    \captionof{figure}{Comparison of offline learning tasks for \swordtask (left) and \pogotask (right) on different map sizes, where \pddlaT denotes the ``zero shot'' results.}
    \label{fig:offline_tasks}
\end{table}

A key advantage of planning lies in its capacity to generalize across varying numbers of objects. 
In contrast, offline \rl and \il algorithms do not inherently possess the same capability.\footnote{Although architectures such as Graph Neural Networks offer a limited form of generalization, achieving this requires non-trivial engineering efforts and often results in degraded performance when the number of objects changes~\citep{munikoti2023challenges}.}
To demonstrate the ability of \pddla to generalize without retraining, we conducted a ``zero-shot'' experiment in which 
training was done on trajectories for $6\times 6$ maps, 
but the test problems were on the larger $10\times 10$ and $15\times 15$ maps. 
The results of this experiment are presented in Figure~\ref{fig:offline_tasks}, 
which plots the success rate as a function of the number of given expert trajectories. 
Each subfigure corresponds to a different model and/or map size. 
The average across the 5 folds is represented by a line, and the corresponding standard deviation is visualized as a shaded area with low opacity around each line.

The results of our zero-shot experiment are shown in the data series labeled by \pddlaT .
For reference, we also plot the results for \bc and \pddla that are trained with trajectories for maps of the same size as the test maps. Note that for the $6\times 6$ map \pddla and \pddlaT are exactly the same algorithm. 
As can be seen in the results for the larger maps ($10\times 10$ and $15\times 15$), 
the \pddla version trained on smaller maps (\pddlaT) performs nearly equivalently to \pddla, 
which was trained directly on larger maps. 
This indicates that our model-based approach is able to generalize well to larger problems while only training on smaller problems. 
An additional trend that can be observed in these results is that for the simpler \swordtask, \bc is actually able to learn faster than both \pddla and \pddlaT. 
For example, given only 40 trajectories to train on, \bc can surpass a success rate of 0.8 in all map sizes for the \swordtask, while neither \pddla nor \pddlaT can do so. 
However, after training on 100 trajectories, both \pddla and \pddlaT are able to match and, for larger maps, surpass the success rate of \bc. Considering also the results in Table~\ref{tab:offline_results}, we see that while \bc is able to quickly learn to solve the easy task, it struggles with the harder tasks. The results for the \pogotask support this conclusion, where in this harder task \bc is significantly outmatched by \pddla and \pddlaT.

\section{Online Learning of Numeric Action Models}
In the online learning setting, no expert trajectories are given, and we control an agent that interacts with the environment in a sequence of \emph{episodes}. 
In every episode, the agent starts in some initial state $s$ and is given a goal to achieve $g$. 
It then performs actions in the environment until it either achieves its goal or has reached some other termination condition, e.g., reaching a terminal state after performing a predefined number of actions. 
Note that in the online learning setting, planning and learning can be interleaved as more observations are added when the agent performs more actions in the domain. 
The primary metrics of interest in this context are the number of episodes in which the agent successfully achieves the goal and the number of steps required until a goal is found in each episode. 
These metrics can be viewed as a form of maximizing cumulative reward in our context.

This problem setting aligns with the conventional \rl setting: an agent engages in actions within the environment, receives observations, and adapts its behavior over time. 
Thus, with the \mcraft to AI gym translation mechanism mentioned above, we can use any standard \rl techniques, e.g., \dqn and \ppo, to solve this online learning problem in our \mcraft benchmark. 

\subsection{The RAMP Hybrid Strategy}
\label{sec:hybrid}

\begin{figure}
    \centering
    \includegraphics[width=\linewidth]{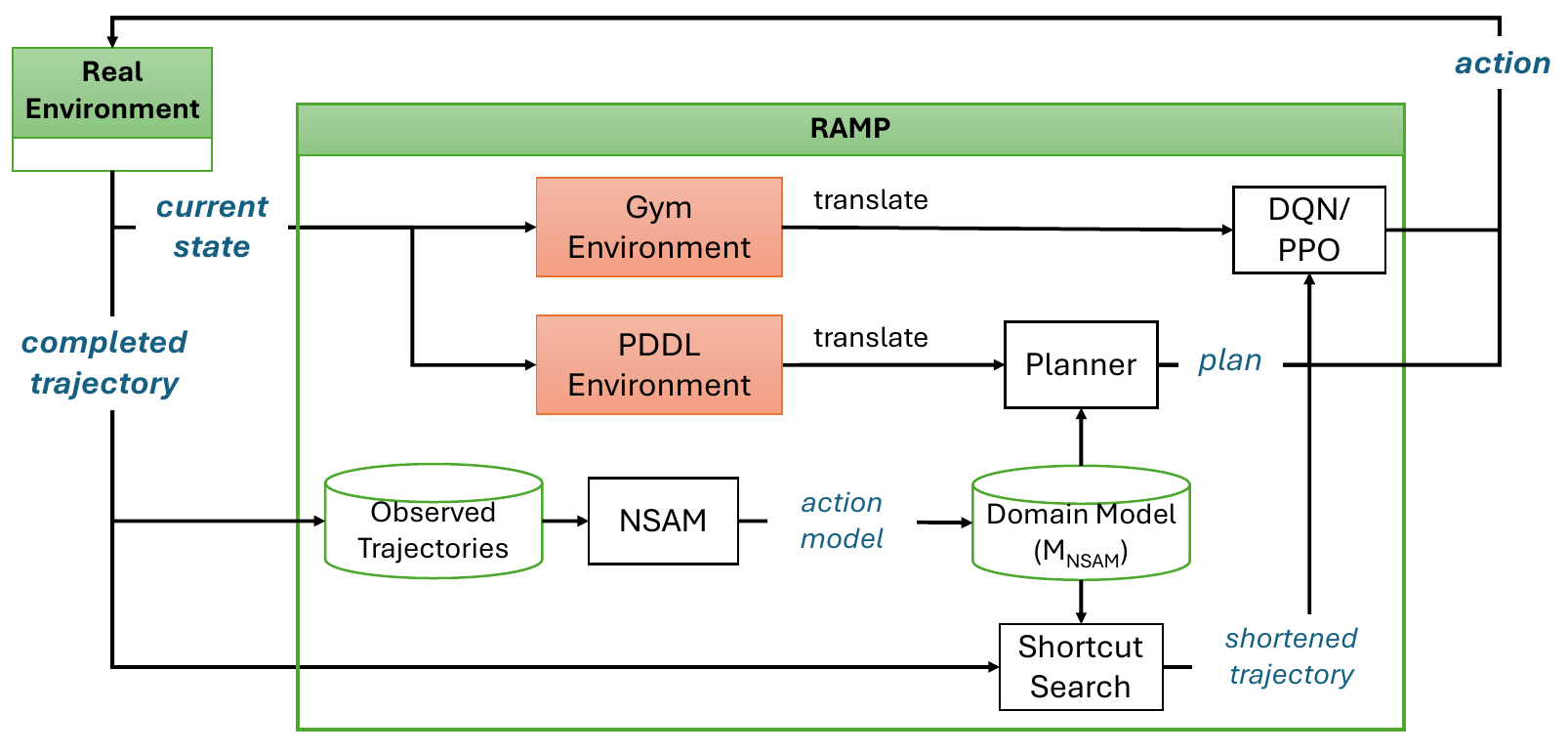}
    \caption{A high-level diagram of the \hybrid hybrid strategy.}
    \label{fig:ramp-diagram}
\end{figure}

The model-based solution described above (\pddla) for the offline learning setting cannot be directly applied in our online problem setting since it lacks a mechanism for collecting trajectories, which is crucial in an online environment. To address this, we introduce a novel hybrid strategy that we call \hybrid. 
\hybrid integrates three components: an \rl algorithm, the \nsam algorithm, and a numeric planner. 
It maintains a set $\mathcal{T}$ of observed trajectories 
and an incumbent learned domain model, denoted \mnsam. 
Both are initialized to be empty. 
At the beginning of every episode, \hybrid attempts to find a plan with the planner using \mnsam. 
If a plan is found, we let the agent execute this plan.\footnote{In the first episode, \mnsam is empty, and thus, the planner cannot run. We consider this as if the planner attempted to find a plan and failed.} 
Otherwise, \hybrid uses the \rl algorithm to choose which actions to perform in this episode. 
At the end of every episode, we add the resulting trajectory to $\mathcal{T}$. 
If the \rl algorithm was able to reach the goal, we perform several additional steps. 
First, \hybrid runs \nsam on the updated set of trajectories to update \mnsam. 
Then, we attempt to run the numeric planner on the same problem instance using the update \mnsam. 
Otherwise, we search for shortcuts in the trajectory found by the \rl algorithm. 
The plan or the resulting shortened trajectory is then passed to the \rl algorithm as an additional training example.
This allows the \rl algorithm to further improve its policy using higher-quality examples. 
Figure~\ref{fig:ramp-diagram} provides a high-level illustration of \hybrid . 
Next, we describe in more detail the different components of \hybrid.

\paragraph{Searching for shortcuts in an existing trajectory} 
The input to the search for shortcuts process is a trajectory $T=(s_0,a_1,s_1,\ldots,a_n,s_n)$, which was generated by running the \rl algorithm in an episode until the agent has reached the goal. 
First, we remove loops in $T$. 
That is, we iterate over the states visited by the agent in the episode and remove every sequence of actions in the plan that start and end in the same state. 
Then, we search for additional shortcuts by searching for sequences of actions in $T$ that can be replaced by a single action according to \mnsam. 
When such a sequence is found, we replace it in $T$ with the corresponding single action. 
There are multiple ways to search for such sequences of actions. 
In our implementation, we performed the following simple procedure to do so. 
First, we set \sto to be the last state in $T$, 
$s$ to be the state before it, and 
\sfrom the state before $s$. I.e., $\sfrom,s,\sto$ are initialized as the three last (consecutive) states in $T$. We denote by $|\actions(T)|$ the number of actions in $T$. 
Thus, $\sto=s_{|\actions(T)|}$, $s=s_{|\actions(T)|-1}$, and $\sfrom=s_{|\actions(T)|-2}$. 
Next, we check if, according to \mnsam, there is an action $a'$ that is applicable in $\sfrom$, and if we apply $a'$ in \sfrom, we directly reach \sto. If this is the case, 
we replace in $T$ the sequence of states and actions starting from \sfrom and ending in \sto with the triplet $(\sfrom, a', \sto)$.
Otherwise, we move \sto one step back to its preceding state and repeat the process.
The procedure terminates when the index of \sto reaches 1.
Algorithm~\ref{alg:shortcuts} lists the pseudo-code for our shortcut search process. Note that its runtime complexity is linear in $T$. More involved shortcut search processes may require higher computational complexity.

\begin{algorithm}
\DontPrintSemicolon
\SetKwInOut{Input}{Input}
\Input{A trajectory $T=(s_0,a_1,s_1,\ldots,a_n,s_n)$}
Remove every loop from $T$ \\
$i \gets 0$; $\sto\gets s_{|\actions(T)|}$ \\ 
\While{$|\actions(T)|-i-2\geq 0$ is }{
    $\sfrom \gets s_{|\actions(T)|-i-2}$ \\
    \eIf{$\exists a'\in$\mnsam where $a'$ is applicable in $\sfrom$\ and $a'(\sfrom) = \sto$}{
           Replace in $T$ the sequence $(\sfrom,\ldots, \sto)$ with $(\sfrom,a',\sto)$
        }{
            $i\gets i+1$; 
            $\sto\gets s_{|\actions(T)|-i}$ \\ 
        }
}
Return $T$ \\
\caption{Shortcut Search in Action Sequences}
\label{alg:shortcuts}
\end{algorithm}

\noindent The following example illustrates our search for shortcuts process. 
\begin{example}    
\label{ex:shortcut}
Consider a $3 \times 3$ grid where the agent is a mobile robot that has eight possible move actions: 
\textsl{move\_up, move\_down, move\_left, move\_right, move\_up\_right, move\_up\_left, move\_down\_right, move\_down\_left}.
Each action moves the robot in the corresponding direction, provided it remains within the grid boundaries. 
A state in this environment is the location of the agent in the grid, i.e., $(x, y)$. 
Now assume that after running \hybrid for several episodes in this environment, \nsam learned a domain model \mnsam that includes the true preconditions and effects for the following actions:
\textsl{move\_up, move\_left, move\_right, move\_up\_right, move\_up\_left}.
Then, we begin a new episode in which the agent starts in $(0,0)$, corresponds to the bottom-left corner of the grid, and its goal is to reach the top-right corner, $(2,2)$. 
\hybrid tries to find a plan for this episode
by running a planner using the \mnsam domain model. 
The planner, however, fails to find a plan within a reasonable time limit. This may occur since planning is, in general, intractable, even when we have the true domain model.  
So, \hybrid runs \ppo and moves the agent accordingly. Eventually, the agent reaches the goal.  
Let $T_{new}$ be the resulting trajectory, illustrated in Figure~\ref{tab:example-shortcut}(a): 
\[
\begin{aligned}
T_{new} = \{ & (0,0), \texttt{move\_up}, (0,1), \texttt{move\_right}, (1,1), \texttt{move\_left}, (0,1),\\
                      & \texttt{move\_right}, (1,1), \texttt{move\_up}, (1,2), \texttt{move\_right}, (2,2) \}
\end{aligned}
\]
First, we run \nsam again with the added new trajectory $T_{new}$, updating \mnsam accordingly. 
Next, we remove loops. In this example, the sequence $((0,1), \texttt{move\_right}, (1,1), \texttt{move\_left}, (0,1))$ forms a loop so we remove it.
The resulting trajectory, illustrated in Figure~\ref{tab:example-shortcut}(b), is:
\[
T' = \{ (0,0), \texttt{move\_up}, (0,1), \texttt{move\_right}, (1,1), \texttt{move\_up}, (1,2), \texttt{move\_right}, (2,2) \}
\]
Next, we attempt to further shorten the corresponding plan by removing sequences of actions that can be replaced by single actions according to \mnsam. 
We can see that according to \mnsam 
move\_up\_right is applicable in $(1,1)$ and $move\_up\_right((1,1))=(2,2)$. 
Thus, we modify $T'$ as shown in Figure~\ref{tab:example-shortcut}(c): 
\[
T'' = \{ (0,0), \texttt{move\_up}, (0,1), \texttt{move\_right}, (1,1), \texttt{move\_up\_right}, (2,2) \}
\]
Next, we continue searching for an action $a''$ that is applicable in $(0,1)$ and $a''((0,1))=(2,2)$. 
No such action exists, so we search for an action $a''$ that is applicable in $(0,0)$ and $a''((0,0))=(1,1)$. According to \mnsam, such an action exists --- move\_up\_right. 
Thus, we modify $T''$ as shown in Figure~\ref{tab:example-shortcut}(d): 
\[
T''' = \{ (0,0), \texttt{move\_up\_right}, (1,1), \texttt{move\_up\_right}, (2,2) \}
\]
No further shortcuts exist, so the shortened trajectory $T'''$ is returned. 
\end{example}

\begin{table}
    \centering
    \begin{tabular}{c|c|c|c}
    \begin{tikzpicture}
    \draw[thick] (0,0) grid (3,3);    

    \node[agent] at (0.5,0.5) {};
    \node[agent-head] at (0.5,0.75) {};

    \draw[thick, red] (2.4,2.2) -- (2.4,2.8); 
    \draw[thick, red, fill=red] (2.4,2.8) -- (2.8,2.65) -- (2.4,2.5) -- cycle; 

    \draw[thick, blue, ->] (0.5,0.5) -- (0.5,1.45);
    \draw[thick, blue, ->] (0.55,1.45) .. controls (1,1) .. (1.45,1.45);
    \draw[thick, blue, ->] (1.45,1.55) .. controls (1,2) .. (0.55,1.55);
    \draw[thick, blue, ->] (1.5,1.5) -- (1.5,2.45);
    \draw[thick, blue, ->] (1.5,2.5) -- (2.35,2.5);
\end{tikzpicture} &  

\begin{tikzpicture}
    \draw[thick] (0,0) grid (3,3);    

    \node[agent] at (0.5,0.5) {};
    \node[agent-head] at (0.5,0.75) {};

    \draw[thick, red] (2.4,2.2) -- (2.4,2.8); 
    \draw[thick, red, fill=red] (2.4,2.8) -- (2.8,2.65) -- (2.4,2.5) -- cycle; 

    \draw[thick, blue, ->] (0.5,0.5) -- (0.5,1.45);
    \draw[thick, blue, ->] (0.5,1.5) -- (1.45,1.5);
    \draw[thick, blue, ->] (1.5,1.55) -- (1.5,2.45);
    \draw[thick, blue, ->] (1.5,2.5) -- (2.35,2.5);
\end{tikzpicture} & 
\begin{tikzpicture}
    \draw[thick] (0,0) grid (3,3);    

    \node[agent] at (0.5,0.5) {};
    \node[agent-head] at (0.5,0.75) {};

    \draw[thick, red] (2.4,2.2) -- (2.4,2.8); 
    \draw[thick, red, fill=red] (2.4,2.8) -- (2.8,2.65) -- (2.4,2.5) -- cycle; 

    \draw[thick, blue, ->] (0.5,0.5) -- (0.5,1.45);
    \draw[thick, blue, ->] (0.5,1.5) -- (1.45,1.5);
    \draw[thick, blue, ->] (1.5,1.5) -- (2.35,2.35);
\end{tikzpicture}         &
\begin{tikzpicture}
    \draw[thick] (0,0) grid (3,3);    

    \node[agent] at (0.5,0.5) {};
    \node[agent-head] at (0.5,0.75) {};

    \draw[thick, red] (2.4,2.2) -- (2.4,2.8); 
    \draw[thick, red, fill=red] (2.4,2.8) -- (2.8,2.65) -- (2.4,2.5) -- cycle; 

    \draw[thick, blue, ->] (0.65,0.65) -- (1.45,1.45);
    \draw[thick, blue, ->] (1.5,1.5) -- (2.35,2.35);
\end{tikzpicture}
         \\
    (a) Initial plan &
    (b) Removing loops &
    (c) Shortcut \#1 &
    (d) Shortcut \#2          
    \end{tabular}
    
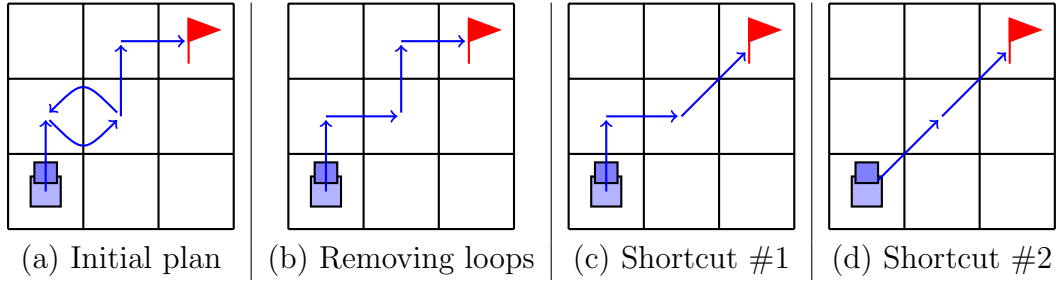
\captionof{figure}{An illustration of Example~\ref{ex:shortcut} of the search for shortcut process.}
    \label{tab:example-shortcut}
\end{table}

\paragraph{Feedback loop to the \rl algorithm}
Every trajectory executed by following a plan returned by the planner is also given to the \rl algorithm as an additional episode for training. 
In addition, if applying the search for shortcuts on an executed trajectory was able to generate a shorter trajectory, then the resulting trajectory is also given to the \rl algorithm as an additional episode for training. 
Note that the shortened trajectory is a valid trajectory because the domains are deterministic, and thus loops are redundant, and due to \nsam's safety property.

This integration of model-based and model-free algorithms in our hybrid strategy establishes a symbiotic relationship between them. 
The model-free \rl algorithm acts as a methodological tool to solve problems when planning fails, as well as to gather information in a goal-oriented manner, leveraging its inherent ability to balance exploration and exploitation. 
Simultaneously, the \rl algorithm benefits from this partnership since trajectories created by the model-based algorithm (either by the planner or by the shortcut search process) often tend to represent more efficient ways to solve the task at hand. Thus, these trajectories provide high-quality data that can improve sample efficiency and stabilize the learning process. 
Our experimental results show that this integrative approach of automated planning and \rl can significantly surpass standard \rl techniques.

\paragraph{Details on using PPO in \hybrid}
After experimenting with several \rl algorithms, we used in our experiments \ppo. 
\ppo is an on-policy algorithm, allowing it to learn from trajectories created by the planner forcing the algorithm to sample the action selected by the planner for these states. 
However, PPO uses a clipping mechanism designed to constrain policy updates by preventing the probability ratio between the new and old policies from deviating too far from 1. When the agent is forced to follow an expert plan, its own learned policy may assign a low probability to the dictated actions, creating a significant discrepancy between the executed actions and those preferred by the current policy. This can cause the importance-weighted policy ratio to frequently fall outside the clipping range, effectively nullifying the gradient update and slowing down learning. 
As a result, in some cases, PPO may struggle to meaningfully adjust its policy, particularly if the expert actions are substantially different from what it would naturally choose. 
To address this issue, we adopt an approach for masking invalid actions~\citep{HuangO22}. Specifically, we treat the expert action in each state as the only valid action and mask out all others. This ensures that the logits of actions that do not conform with the plan are zeroed out, preventing them from influencing the policy update. As a result, the gradient for these actions is not eliminated, ensuring that the update is not affected by PPO's clipping mechanism.

\subsection{Experimental Setup}
We conducted an experimental evaluation to compare 
\ppo with action masking and \hybrid. 
We used the same \mcraft domain, tasks, and maps described in Section~\ref{sec:setup-offline}. For each task and map size, we created 50 different problem instances. 
Then, we run each of the evaluated algorithms on each problem instance for a fixed number of steps, denoted $B_i$,  before passing to the next problem instance. Additionally,
in every episode, we run the evaluated algorithm on the current problem instance until either the agent reaches the goal or it has performed a fixed number of steps, denoted $B_e$. 
We set $B_i$ and $B_e$ differently for the \swordtask and \pogotask since they differ in their difficulty and required plan length.  
For the \swordtask task, $B_i=800$ and $B_e=200$, while for the \pogotask task, $B_i=6,000$ and $B_e=1,500$. 
Note that $B_i>B_e$, and thus each algorithm will run multiple episodes on the same problem instance before passing to the next problem instance. 
This allows the evaluated algorithms to learn better how to solve problems in this non-trivial environment. 
In each experiment, we measured how many problems each algorithm was able to solve and the minimum plan length it found for each problem instance. 
Since \ppo is a stochastic algorithm, we repeated every experiment five times, each time using a different random seed.

\paragraph{Evaluated algorithms and implementation details}

Preliminary experiments with \dqn and \qrdqn instead of \ppo showed that it consistently failed to solve the \bothtasks tasks, 
despite exhaustive efforts in hyperparameter tuning and experimentation with various network architectures. 
This was also observed in our offline learning results.
Therefore, we omit the \dqn and \qrdqn results from our presentation. 
The \ppo implementation we used is from the \texttt{stable\_baselines3}\footref{ft:sb3} python library. 
To leverage the deterministic nature of the domain, we used \ppo with action masking~\citep{TangLCY20} instead of regular \ppo. This version of \ppo prevents executing actions that were previously attempted in the current state and deemed inapplicable.

After some preliminary experiments and fine-tuning, we chose for \ppo a fully connected neural network architecture with two layers, each having 64 units and a $tanh$ activation function. 
This network was trained with the following configuration: an entropy coefficient of 0.01, a discount factor of 0.999, a value function coefficient of 0.65, and a maximum gradient clipping of 1.0.

For our hybrid strategy (\hybrid), we used \ppo as the \rl algorithm and MetricFF~\citep{hoffmann2003metric} as the planner. 
\noindent We also run the following simplified versions of \hybrid: 
\begin{itemize}
    \item \textbf{\hybrido .} This version of \hybrid never runs a planner. It only attempts to find shortcuts, as explained in Section~\ref{sec:hybrid}. 
    \item \textbf{\hybridl .} This version of \hybrid does not run a planner or \nsam. It only attempts to find shortcuts by removing loops. 
\end{itemize}
These versions serve for ablation study purposes. Note that \hybridl does not use any model-based reasoning.

\paragraph{Avoiding planning when redundant}
Calling the planner can be very costly in terms of runtime. 
Therefore, in our implementation we avoided running the planner in cases where we know it will not provide any benefit. Specifically, we did not call the planner in the following cases. 
\begin{itemize}
    \item \textbf{In the initial episode, and every subsequent episode until a goal has been reached for the first time.}
    These calls are redundant since in both \mcraft tasks, reaching a goal involves performing a special action (\swordtask or \pogotask) that is only used when reaching a goal state. Thus, until performing this action at least once, there is no hope of finding a plan. 
    \item \textbf{In problem instances where the planner was already called, and \mnsam has not been updated since.} These calls are redundant because the planner used in our experiments is deterministic. 
    Therefore, calling the planner on the same problem instance with the same action model will return the same outcome. 
\end{itemize}

\paragraph{Performance Metrics}
We consider two main metrics in these experiments: \emph{success rate} and \emph{minimum plan length}. 
The \emph{success rate} here is the ratio of problem instances that the algorithm was able to solve. Since each algorithm performs multiple episodes on the same problem instance, it is sufficient for at least one of them to reach the goal in order to count this problem instance as solved by the algorithm. 
Similarly, the \emph{minimum plan length} found by algorithm $X$ for problem instance $Y$ is the minimum over the lengths of all plans algorithm $X$ found for problem instance $Y$.

\subsection{Experimental Results}

\begin{table}[ht]
    \centering
    \small
    \begin{tabular}{|c|c|cccc|cccc|}
    \hline
    Map Size & Inst. & \multicolumn{4}{c|}{Wooden Sword} & \multicolumn{4}{c|}{Pogo Stick} \\
    \hline
             &       & PPO & $R_{(-pn)}$ & $R_{(-p)}$ & $R$ & PPO & $R_{(-pn)}$ & $R_{(-p)}$ & $R$ \\
    \hline
    \multirow{5}{*}{6×6} 
         & 10 & 0.96 & \textbf{0.98} & 0.96 & \textbf{0.98} & 0.88 & 0.96 & 0.90 & \textbf{1.00} \\
         & 20 & 0.98 & \textbf{0.99} & 0.98 & \textbf{0.99} & 0.82 & 0.82 & 0.83 & \textbf{0.99} \\
         & 30 & \textbf{0.99} & \textbf{0.99} & \textbf{0.99} & \textbf{0.99} & 0.88 & 0.88 & 0.88 & \textbf{0.99} \\
         & 40 & 0.98 & 0.98 & 0.98 & \textbf{0.99} & 0.91 & 0.91 & 0.91 & \textbf{1.00} \\
         & 50 & 0.97 & 0.98 & 0.97 & \textbf{0.99} & 0.92 & 0.92 & 0.93 & \textbf{1.00} \\
    \hline
    \multirow{5}{*}{10×10} 
         & 10 & 0.92 & \textbf{1.00} & \textbf{1.00} & 0.98 & 0.76 & 0.76 & 0.86 & \textbf{0.88} \\
         & 20 & 0.93 & \textbf{1.00} & \textbf{1.00} & 0.99 & 0.85 & 0.87 & 0.91 & \textbf{0.92} \\
         & 30 & 0.93 & \textbf{0.98} & \textbf{0.98} & \textbf{0.98} & 0.84 & 0.90 & \textbf{0.93} & \textbf{0.93} \\
         & 40 & 0.93 & 0.97 & 0.98 & \textbf{0.99} & 0.85 & 0.93 & 0.94 & \textbf{0.95} \\
         & 50 & 0.94 & 0.97 & 0.98 & \textbf{0.99} & 0.86 & 0.93 & 0.94 & \textbf{0.95} \\
    \hline
    \multirow{5}{*}{15×15} 
         & 10 & 0.82 & 0.84 & \textbf{0.90} & 0.84 & 0.42 & 0.46 & 0.52 & \textbf{0.56} \\
         & 20 & 0.85 & 0.88 & \textbf{0.90} & 0.88 & 0.51 & 0.47 & 0.63 & \textbf{0.67} \\
         & 30 & 0.83 & 0.89 & \textbf{0.90} & \textbf{0.90} & 0.59 & 0.53 & 0.75 & \textbf{0.78} \\
         & 40 & 0.86 & 0.92 & \textbf{0.93} & \textbf{0.93} & 0.62 & 0.52 & 0.78 & \textbf{0.81} \\
         & 50 & 0.88 & \textbf{0.94} & \textbf{0.94} & \textbf{0.94} & 0.68 & 0.56 & 0.82 & \textbf{0.84} \\
    \hline
    \end{tabular}
    \caption{Comparison of online learning tasks for \swordtask and \pogotask on different map sizes, focusing on average success rate (higher is better).
    }
    \label{tab:online_tasks-success}
\end{table}

Table~\ref{tab:online_tasks-success} shows the average success rate for every task, algorithm (columns), map size, and number of problem instances the agent has already trained on (rows). 
The columns $R$, $R_{(-p)}$, and $R_{(-pn)}$, correspond to \hybrid, \hybrido, and \hybridl, respectively. 
The best result in each configuration is highlighted in bold. 
These results show that all variants of \hybrid outperform \ppo in all configurations but the simplest one (\swordtask, $6\times 6$). 
For example, in the \pogotask task, map size $15\times 15$, and 50 problem instances, \ppo reached only 0.68 average success rate while \hybrid achieved an average success rate of 0.84.

Our ablation study on the different components of \hybrid reveals surprising results. Only removing loops and providing this back to the \ppo algorithm already yields a significant increase in success rate.
In fact, the benefit of learning an action model, which is used by \hybrid and \hybrido, is only observed in the results of the harder task (\pogotask). 
In the \pogotask results, however, \hybridl performs significantly worse than the full \hybrid, emphasizing the merit of learning a numeric domain model.

\begin{table}[htbp]
    \centering
    \begin{tabular}{c c}
                \begin{minipage}{0.45\textwidth}
            \centering
            \begin{minipage}[b]{\textwidth}
                \includegraphics[width=\textwidth]{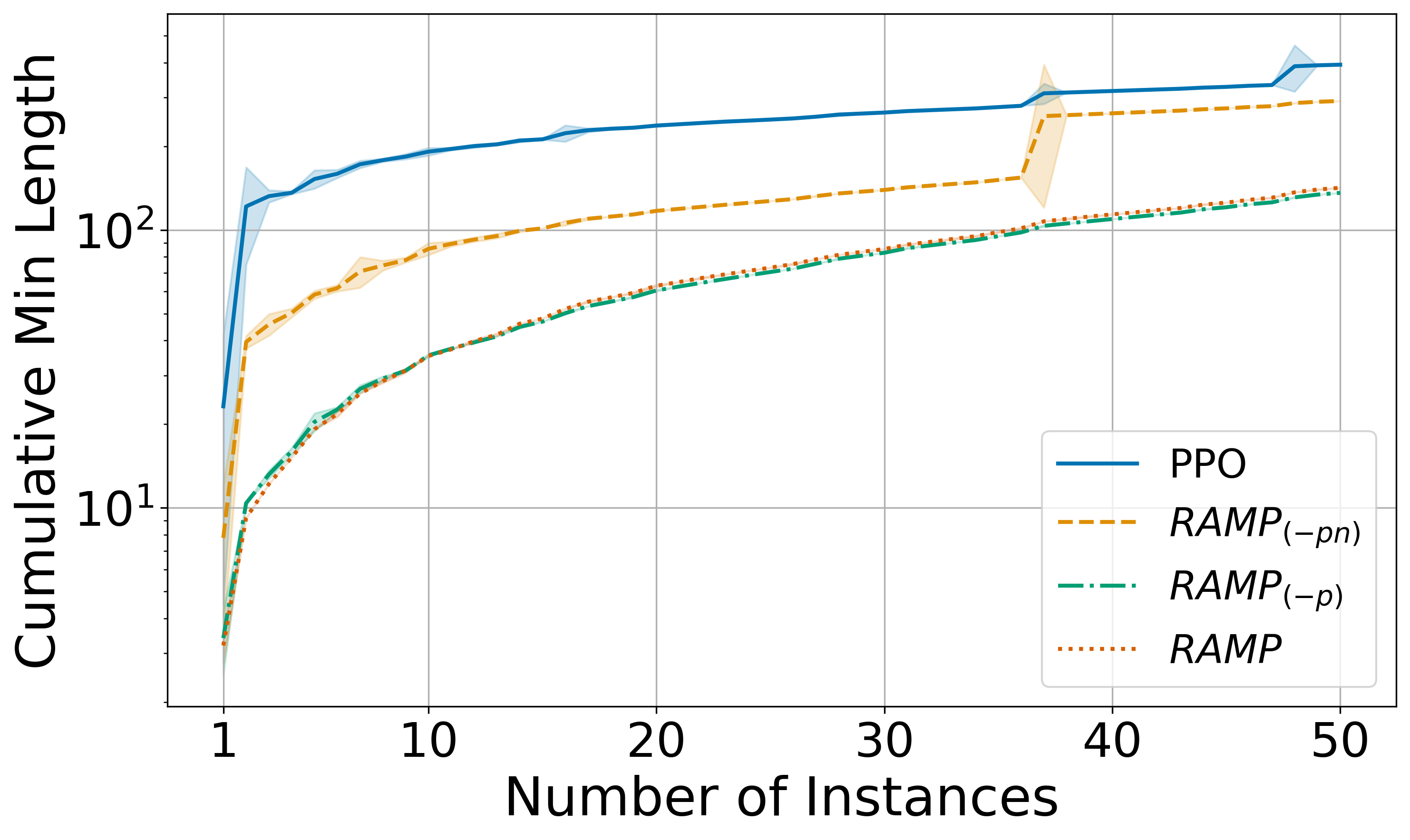}
                \captionof{figure}{\swordtask task on $6\times6$ maps}
                \label{fig:6x6-online-sword-length}
            \end{minipage}
            \medskip
            \begin{minipage}[b]{\textwidth}
                \includegraphics[width=\textwidth]{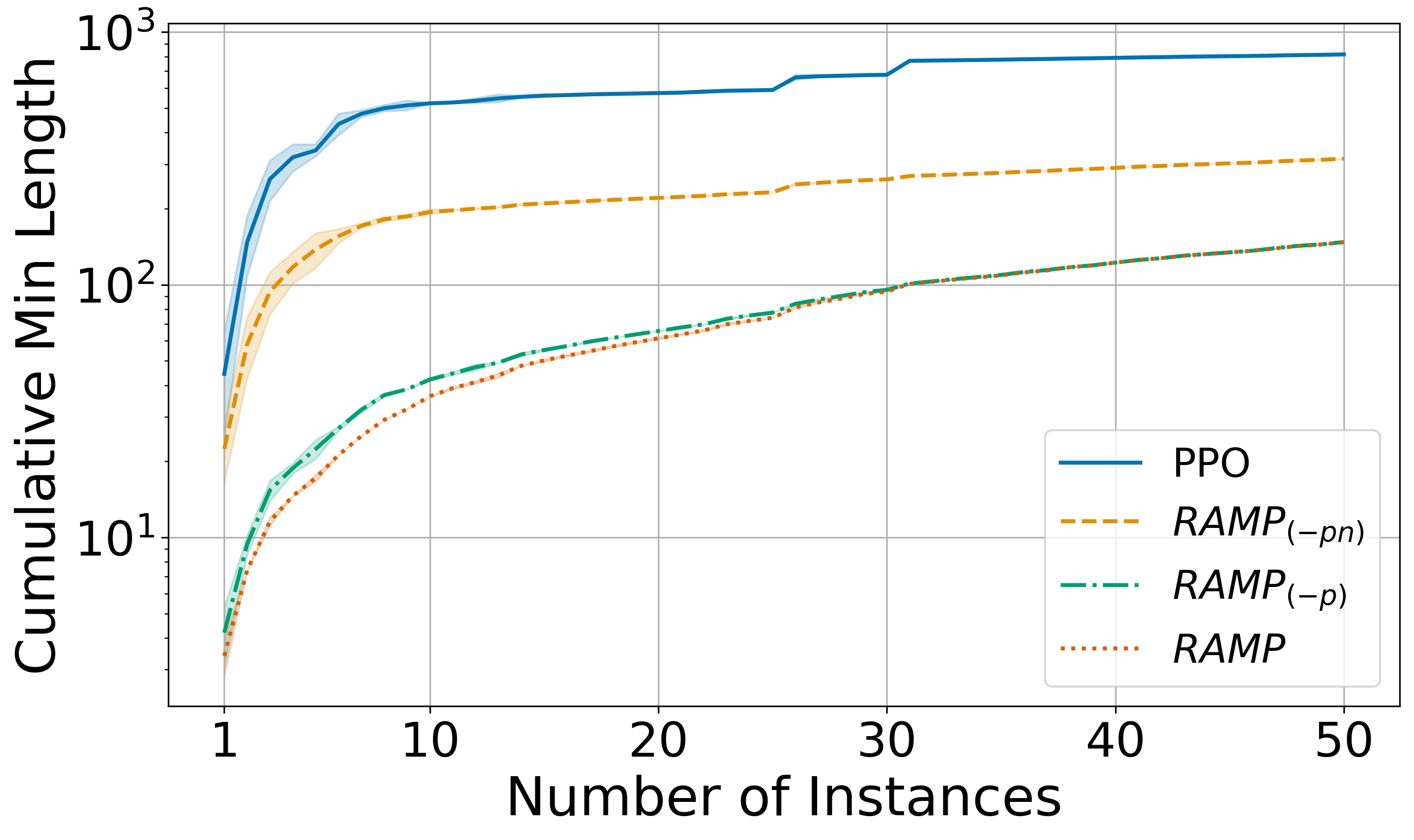}
                \captionof{figure}{\swordtask task on $10\times10$ maps}
                \label{fig:10x10-online-sword-length}
            \end{minipage}
            \medskip
            \begin{minipage}[b]{\textwidth}
                \includegraphics[width=\textwidth]{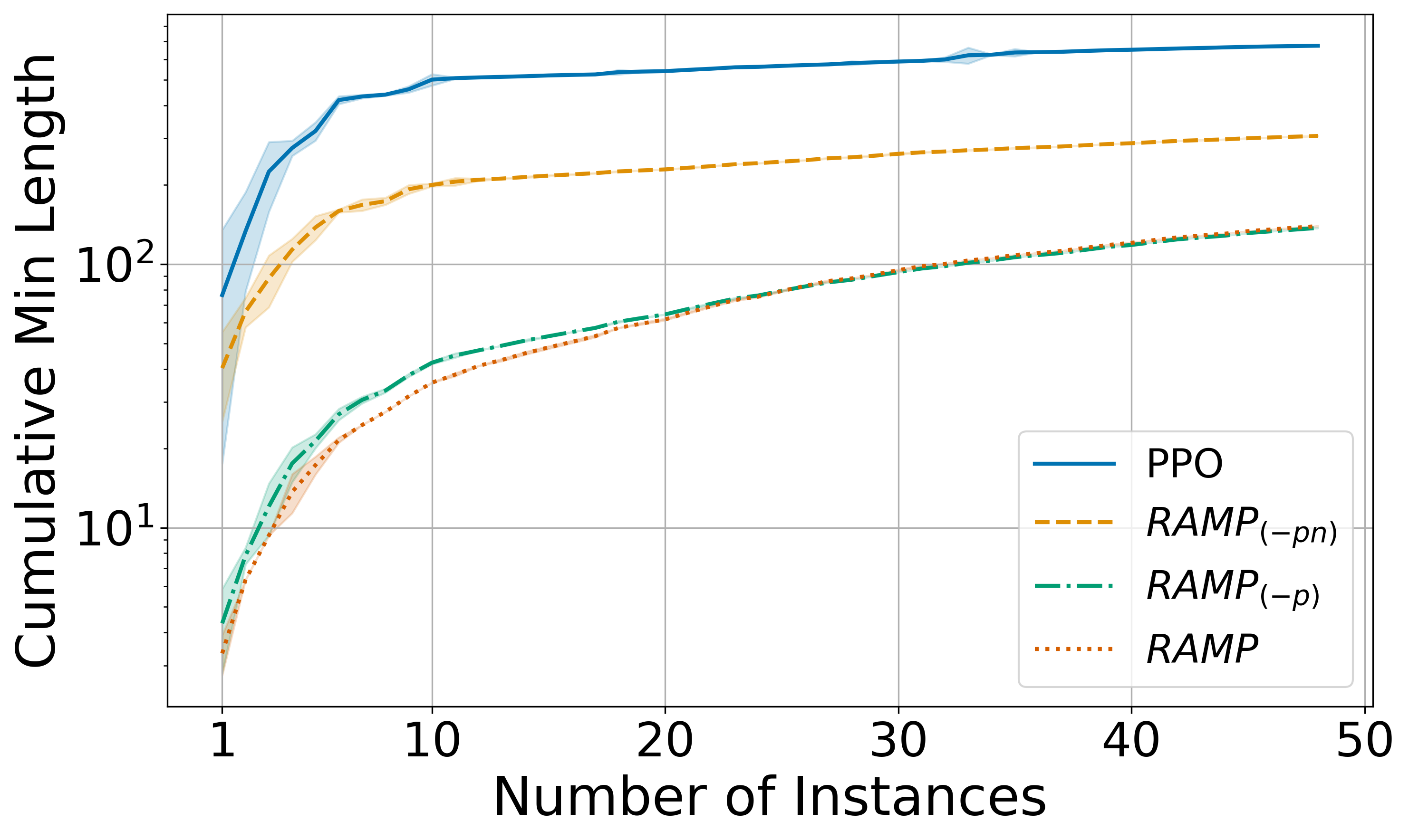}
                \captionof{figure}{\swordtask task on $15\times15$ maps}
                \label{fig:15x15-online-sword-length}
            \end{minipage}
            \label{fig:online_sword-length}
        \end{minipage}
        &
        \begin{minipage}{0.45\textwidth}
            \centering
            \begin{minipage}[b]{\textwidth}
                \includegraphics[width=\textwidth]{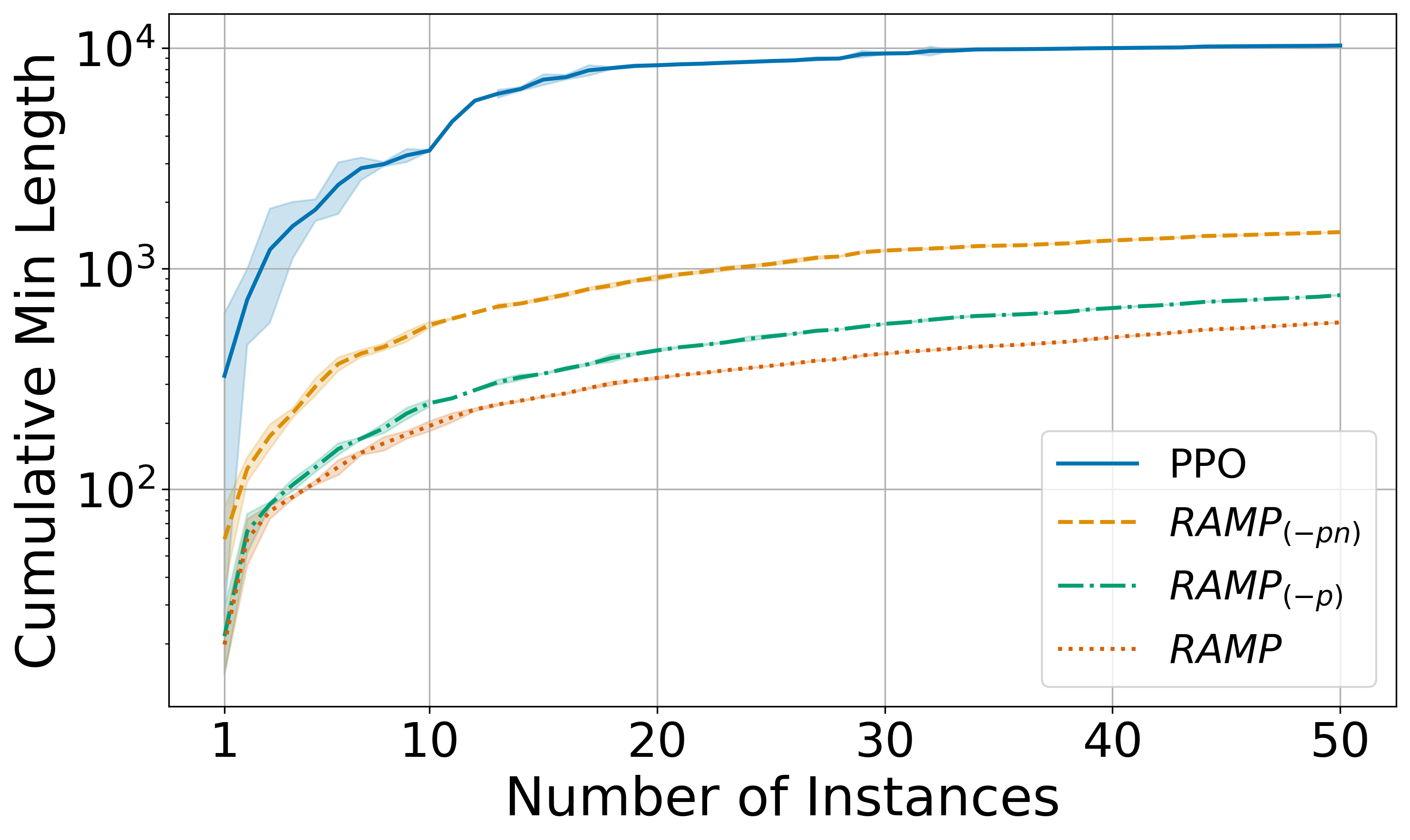}
                \captionof{figure}{\pogotask task on $6\times6$ maps}
                \label{fig:6x6-online-pogo-length}
            \end{minipage}
            \medskip
            \begin{minipage}[b]{\textwidth}
                \includegraphics[width=\textwidth]{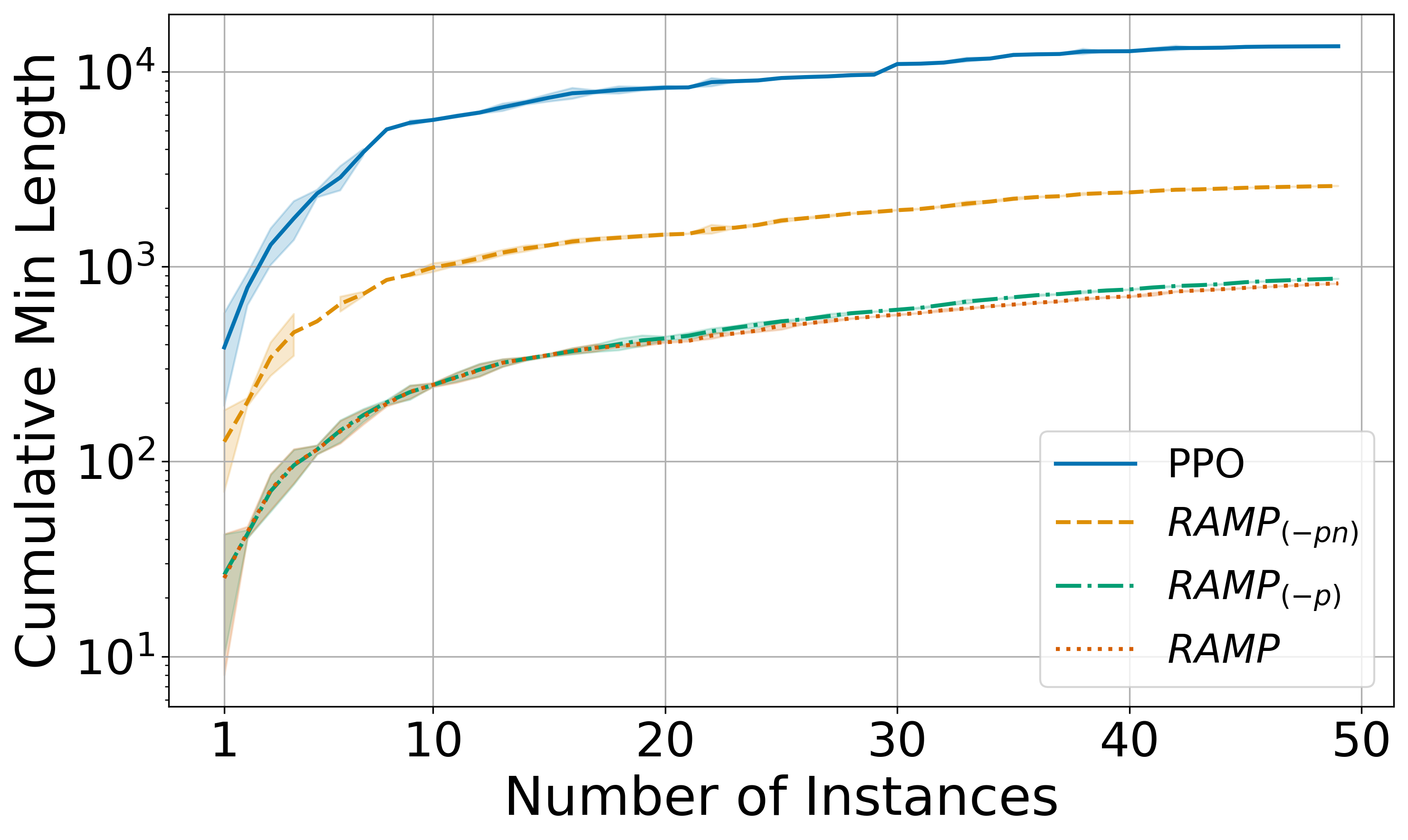}
                \captionof{figure}{\pogotask task on $10\times10$ maps}
                \label{fig:10x10-online-pogo-length}
            \end{minipage}
            \medskip
            \begin{minipage}[b]{\textwidth}
                \includegraphics[width=\textwidth]{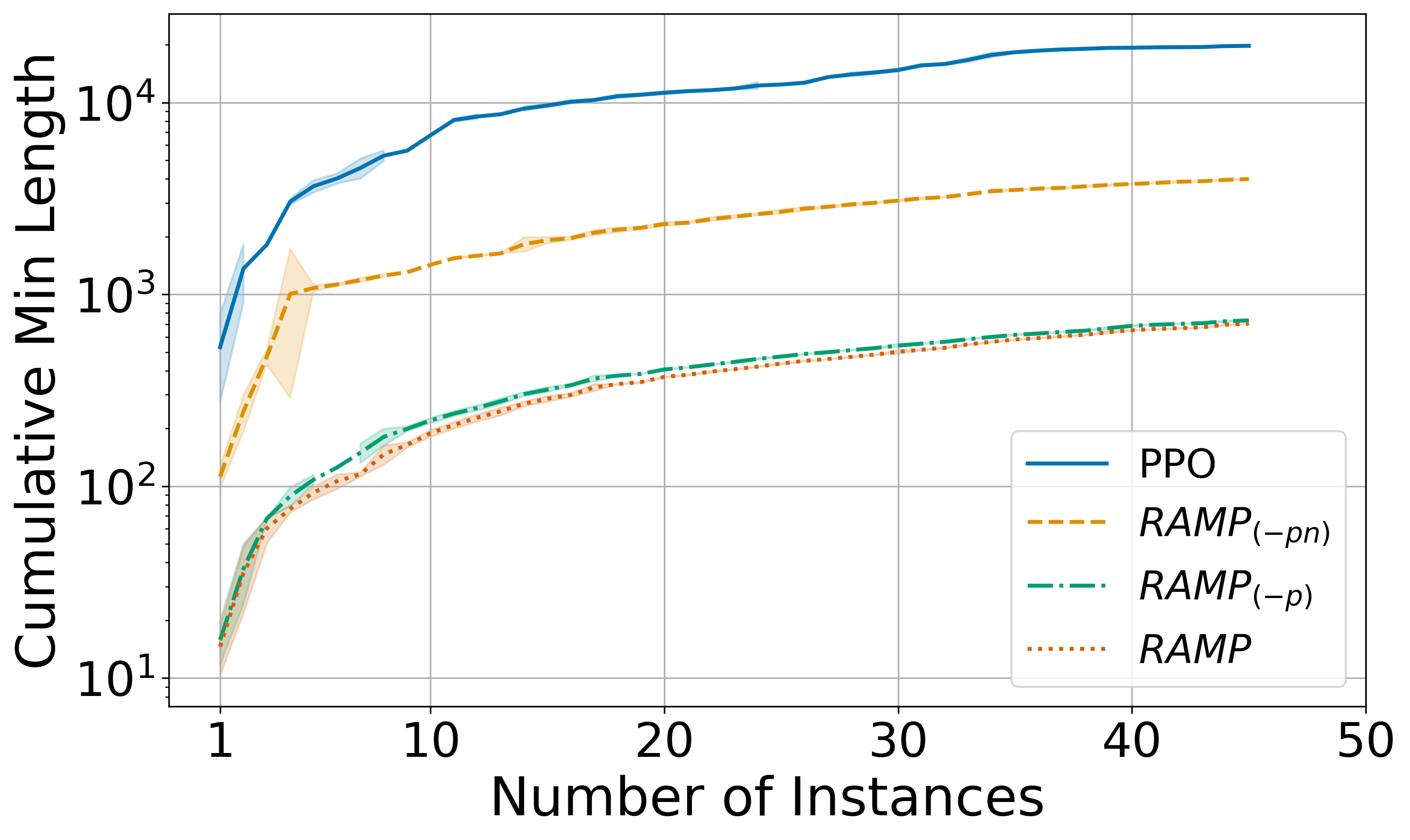}
                \captionof{figure}{\pogotask task on $15\times15$ maps}
                \label{fig:15x15-online-pogo-length}
            \end{minipage}
            \label{fig:online_pogo-length}
        \end{minipage}
    \end{tabular}
    \captionof{figure}{Comparison of online learning tasks for \swordtask and \pogotask on different map sizes, focus on cumulative min length (lower is better).}
    \label{fig:online_tasks-length}
\end{table}

Next, we compare the \emph{minimum plan length} found by each algorithm for every problem instance. 
To avoid the need to set some arbitrarily high value for failed instances, we considered in this analysis only results for problem instances that were solved by all algorithms, in the same random seed. 
Figure~\ref{fig:online_tasks-length} plots on the $y$-axis the \emph{cumulative minimum plan length} in log scale over these results, 
averaged over the relevant random seeds. 
The $x$-axis shows the number of problem instances the agent already trained on. 
By \emph{cumulative minimum plan length} here, we mean we sum the minimum plan lengths found by each algorithm for every problem instance it trained on so far (recall that in our setup, each agent runs multiple episodes on each problem instance).

The results show several clear trends. 
First, all \hybrid variants are able to find significantly shorter plans than the model-free \ppo algorithm. 
In some cases, the plans found by \hybrid are shorter by two orders of magnitude than those found by \ppo. 
Second, in terms of our ablation study, we see that only removing loops, i.e., \hybridl, yields much longer plans than \hybrido and \hybrid, which also use the learned action model (\mnsam). 
This highlights the importance of learning a domain model with \nsam in this environment. 
Lastly, we observe that, surprisingly, the benefit of calling a planner to find shorter plans is relatively minor, and only using the learned action model is sufficient. 
In fact, in some cases, there is almost no clear advantage, in terms of minimum plan length, between \hybrido and \hybrid. For example, see the $6\times 6$ results in \swordtask and the $10\times 10$ and $15\times 15$ results in \pogotask. One explanation for this is that if the problem is too easy, shortcuts are enough to find close-to-optimal plans, while if the problem is too hard the planner may struggle to find plans with the learned action model. 

\section{Related Work}\label{sec:related-work}
In this section, we discuss several lines of research that are related to this work. 

\subsection{Offline Action Model Learning Algorithms}
\label{sec:related-offline-action-model-learning}

FAMA~\citep{aineto2019learning} is an offline action model learning algorithm that can handle missing observations and outputs a classical planning domain model. 
It frames the task of learning an action model as a planning problem, ensuring that the returned action model is consistent with the provided observations.
NOLAM~\citep{Lamanna24} can learn action models even from noisy trajectories. 
LOCM~\citep{cresswell2013acquiring} learns an action model from observed sequences of actions and their signatures, without observing the states in the trajectory.

The Safe Action Model (SAM) learning algorithm~\cite{stern2017efficient,juba2021safe} differs from the above algorithm in that the action model it returns provides a form of ``safety'' guarantee: not only is it consistent with the given observations but it is also guaranteed that every plan created with the learned action model is \emph{sound} with respect to the real, unknown action model (Definition~\ref{def:safe}).
SAM has been extended to support lifted action model representation~\citep{juba2021safe}, partial observability~\citep{le2024learning}, stochastic effects~\citep{juba2022learning}, and conditional effects~\cite{MordochSSJ24}. 
All SAM algorithms require observing all the actions in the given trajectories, and most
also require observing the states. 
The \nsam algorithm used in this work is a member of this family of algorithms. 

All the mentioned above algorithms assume the given observations are symbolic. 
Recent work explored how to learn action models from raw images. 
LatPlan~\citep{asai2018classical} learns propositional action models in the latent space using a variational autoencoder. 
They use the Gumbel-Softmax technique~\citep{jang2017categorical} to convert the continuous output of an autoencoder into categorical variables. 
These categorical variables are used as propositional symbols in a symbolic reasoning system, which in LatPLan's case is a symbolic action model. 
ROSAME-I~\citep{xi2024neuro}, like LatPlan, learns action models from visual inputs. 
Unlike LatPlan, ROSAME-I requires knowing the set of possible propositions and action signatures as input.
ROSAME-I simultaneously learns classifiers for identifying propositions in a given image and infers a lifted, first-order action model defined over the given set of propositions and actions.

\nsam~\cite{mordoch2023learning} and 
PlanMiner~\citep{segura2021discovering} are, to the best of our knowledge, the only algorithms capable of learning action models that include both discrete and continuous preconditions and effects.\footnote{Some other action model learning algorithms are capable of learning the numeric costs or rewards associated with actions~\cite{jin2022creativity}.} 
\nsam makes several simplifying assumptions that allow it to run in polynomial time.
Specifically, it assumes full, noise-free, observability, 
and that preconditions are conjunctions of linear inequalities while effects are conjunctions of linear equations involving numeric state variables. 
Under these assumptions, \nsam provides the same type of safety guarantee as other SAM learning algorithms. 
Unlike \nsam, PlanMiner can learn from noisy observations. 
However, it does not provide any guarantee on the learned action model, and it is unclear whether or not it supports numeric preconditions. 
In addition, it requires solving a symbolic regression problem, which, in general, is computationally intractable. 
For these reasons, we chose \nsam as our main model learning algorithm.

\subsection{Online Action Model Learning Algorithms}
\label{sec:related-online-action-model-learning}
Online action-model learning algorithms 
iteratively learn an incumbent action model and choose the next actions to perform in order to collect observations that enable further refinement of the incumbent action model. 
OLAM~\citep{lamanna2021online} is an online action model learning algorithm that is designed for classical planning domains. It identifies in every iteration an action and a state where trying to execute that action is expected to refine the incumbent action model. 
Then, it uses a planner to find a plan to reach that state and attempts to execute the chosen action. 
GLIB~\citep{chitnis2021glib} follows a similar approach but is designed for stochastic environments, resulting in a Probabilistic PDDL (PPDDL) action model.
QACE~\citep{verma2023autonomous} is an action model learning algorithm that can also query a black-box expert. It outputs a PPDDL action model with the same capabilities as the black-box expert it trained from. 
Karia et al.~\citeyear{karia2023epistemic} extends QACE to address the non-stationarity of the environment, i.e., address cases where the environment dynamics change. QACE+ achieves this by interleaving planning and learning and focusing on learning only the models essential for the tasks at hand.
ILM~\citep{ng2019incremental} employs an explore-exploit strategy: if it reaches a state from which the goal can be achieved, it exploits this state; otherwise, it explores through random walks.
Instead of focusing solely on reaching a specific goal, the agent can take a broader approach by exploring the environment and aiming for an interesting state in it.

Recent works explored integrating \rl and online learning of a symbolic action model~\cite{jin2022creativity,sreedharan2023optimistic}. 
The objective of these works is typically to maximize a cumulative reward metric, in contrast to action model learning algorithms like OLAM and ILM, whose objective is to learn a symbolic action model.
Sreedharan and Katz~\cite{sreedharan2023optimistic} proposed such an algorithm, which we refer to as the SK algorithm. SK begins by initializing an optimistic symbolic model that assumes all actions are applicable in every state (i.e., no preconditions) and the effect of each action includes all grounded predicates. It then employs fast, diverse planners to generate potential paths toward the goal. While these paths are unlikely to be valid, they serve as exploration mechanisms to gather new information. 
Specifically, these plans are executed within the environment, with the outcomes used to train a reward-maximizing policy using Q-learning~\citep{watkins1989learning,watkins1992q}.
This continuous process of exploration and symbolic model refinement is guaranteed to generate a goal-reaching policy. Our hybrid strategy for the online learning setting is somewhat similar to SK. However, SK is only designed for classical (non-numeric) planning, and its applicability to numeric planning remains uncertain. 
SORL~\citep{jin2022creativity} is another online algorithm that integrates RL and learning symbolic action models to maximize the cumulative reward. 
It collects visual observations from the environment and assumes the existence of a mapping function from visual observations to symbolic states. 
SORL iteratively learns and creates 
a symbolic, higher-level action model, 
and a lower-level set of RL policies, referred to as \emph{symbolic options}. 
A planner uses the learned symbolic action model to create a high-level plan, and a meta-controller chooses or creates symbolic options to try to execute the high-level plan, exploring the environment as needed. 

While not explicitly specified, the action model learning algorithm SORL uses is not robust to missing or noisy observations. It assumes that an action's effects can be inferred by the difference between the states observed before and after applying that action and shows no support for numeric preconditions. Numeric effects are supported in a very limited way, only learning which actions increase the reward and by how much. Other numeric aspects, e.g., numeric state variables and preconditions, are not learned.

\subsection{Summary: Action Model Learning Algorithms}
\label{sec:summary-action-model-learning}

\begin{table}[ht]
\centering
\resizebox{\columnwidth}{!}{
\begin{tabular}{|l|c|c|c|c|c|c|c|}
\hline
\textbf{} & \textbf{Input} & \textbf{Numeric} & \textbf{Stochastic} & \textbf{NS} & \textbf{Noisy} & \textbf{Obs.} & \textbf{Online/Offline} \\
\hline
\textbf{FAMA~\citep{aineto2019learning}} & symbolic & No & No & No & No & Partial & Offline \\
\hline
\textbf{LOCM~\citep{cresswell2013acquiring}} & symbolic & No & No & No & No & Only action & Offline \\
\hline
\textbf{OLAM~\citep{lamanna2021online}} & symbolic & No & No & No & No & Yes & Online \\
\hline
\textbf{NOLAM~\citep{Lamanna24}} & symbolic & No & No & No & Yes & Yes & Offline \\
\hline
\textbf{ILM~\citep{ng2019incremental}} & symbolic & No & Yes & Yes & Yes & Yes & Online \\
\hline
\textbf{GLIB~\citep{chitnis2021glib}} & symbolic & No & Yes & No & No & Yes & Online \\
\hline
\textbf{QACE~\citep{verma2023autonomous}} & symbolic & No & Yes & No & No & Yes & Online \\
\hline
\textbf{QACE+~\citep{karia2023epistemic}} & symbolic & No & Yes & Yes & No & Yes & Online \\
\hline
\textbf{SORL~\citep{jin2022creativity}} & visual\footnote{They assumed as input a perfect mapping from visual input to symbolic state.}  & Yes\footnote{The support for numeric planning is limited to only learning how much reward each action adds.} & No & No & - & Yes & Online \\
\hline
\textbf{SAM~\citep{juba2021safe}} & symbolic & No & No & No & No & Yes & Offline \\
\hline
\textbf{NSAM~\citep{mordoch2023learning}} & symbolic & Yes & No & No & No & Yes & Offline \\
\hline
\textbf{PlanMiner~\citep{segura2021discovering}} & symbolic & Yes & No & No & Yes & Partial & Offline \\
\hline
\textbf{SK~\citep{sreedharan2023optimistic}} & symbolic & No & No & No & No & Yes & Online \\
\hline
\textbf{JRK~\citep{james2022autonomous}} & visual & No & Yes\footnote{While they learn a PPDDL model, the experimental results all use a deterministic planner.} & No & - & Yes & Offline \\
\hline
\textbf{LATPLAN~\citep{asai2018classical}} & visual & No & No & No & - & Yes & Offline \\
\hline
\textbf{ROSAME-I~\citep{xi2024neuro}} & visual & No & No & No & - & No\footnote{Note that the first and last states in every trajectory are assumed to be fully obseravable} & Offline \\
\hline
\rowcolor{yellow}
\textbf{\hybrid (our method)} & symbolic & Yes & No & No & No & Yes & Online  \\
\hline
\end{tabular}
}
\caption{Comparison of various action model learning algorithms, based on their support of given problem - numeric inputs, stochasticity, non-stationarity, noisy, observability, and online/offline learning capability.}
\label{tab:methods-comparison}
\end{table}

Table~\ref{tab:methods-comparison} provides an overview of all action model learning algorithms described above. 
Every row represents a model-learning algorithm, and every column represents a property of action model-learning algorithms. 
Column ``Input'' refers to the type of input given to the learning algorithm, namely if it is symbolic or visual. 
Columns ``Numeric'' and ``Stochastic'' refer to whether the underlying environment includes numeric state variables and stochastic effects, respectively. 
Column ``NS'' (non-stationarity) 
refers to whether the dynamics of the underlying environment, i.e., the actions' preconditions and effects, may change during learning. 
Columns ``Noisy'' and ``Obs.'' refer to whether the states and actions in the given observations are noisy and fully observed, respectively. Note that if the ``Input'' is visual, the algorithm can handle noise due to its use of function approximation for image processing. This is indicated by ``-'' in Table~\ref{tab:methods-comparison}. 
The ``Online/Offline'' column refers to whether the learning algorithm is an online algorithm or an offline algorithm. 
As can be seen, the \hybrid hybrid strategy we propose in this work is the only online learning algorithm that supports learning a numeric action model.

\subsection{AI Agents for Minecraft}
\label{sec:related-minecraft}
In our work, \mcraft is merely a benchmark for the evaluation of the proposed models and algorithms. 
However, many prior works have focused on the problem of solving various \mcraft tasks. 
Indeed, both Planning and \rl have been used to design AI agents for \mcraft. 
We provide a brief review of these efforts below.

\citet{roberts2017automated} developed an AI \mcraft agent based on automated planning. They designed a custom PDDL+~\cite{fox2002pddl+} domain model for the \mcraft task. Their agent was able to achieve this task by using a PDDL+ planner and a goal reasoner. Adapting their PDDL+ domain to different tasks is not trivial. 
\citet{wichlacz2019construction} used PDDL modeling to solve complex construction tasks in \mcraft. 
They modeled house-construction tasks as classical and as Hierarchical Task Network (HTN)~\citep{georgievski2015htn} planning problems. 
They observed that even simple tasks pose challenges for classical planners as the size of the world increases. 
In contrast, the HTN planner scaled better but was too coupled with specific tasks.

\citet{james2022autonomous} proposed an algorithm for learning a PPDDL domain from visual input in \mcraft. We refer to their algorithm as the JRK algorithm. 
JRK processes the visual input to identify objects and learn a grounded PPDDL domain. 
Then, it learns a lifted PPDDL domain by grouping objects into types based on the similarity of their preconditions and effects. 
The domain they learned is purely propositional, while for the \mcraft tasks we require domain models that include both propositions and numeric state variables.

Hierarchical Deep Reinforcement Learning Network (H-DRLN)~\citep{tessler2017deep}  
is commonly used in successful MineRL agents. 
When using H-DRLN, the agent continuously learns multiple policies and adapts to new challenges within the game. 
The H-DRLN leverages a deep neural network to model the policy and value functions, resulting in high effectiveness across a variety of \mcraft tasks such as navigation, mining, and combat. Despite its success, this approach requires tens to hundreds of thousands of steps for single, non-hierarchical tasks, while more complex, multi-task hierarchical setups may demand up to several million steps to achieve success. 
Thus, it is less suitable for our context.

Recent work explored how to use Large Language Models (LLMs) to solve \mcraft tasks. 
The VOYAGER algorithm~\citep{wang2024voyager} is an example of an LLM-based \mcraft agent. It comprises three key components: an automatic curriculum for exploration tasks, a skill library for complex behaviors, and an iterative prompting mechanism for generating executable code using JavaScript and the Mineflayer API~\citep{PrismarineJS2013}. It engages with GPT-4 via text prompts, refining generated code based on environmental feedback to successfully complete tasks. Skills accumulated in the library facilitate behavior reuse and unlock new capabilities. 
Using VOYGER has several disadvantages. 
First, it requires many calls to GPT-4, which may be limited and costly. 
Second, it may suffer from \emph{hallucinations}. 
In our case, this means it may ask the agent to perform non-existing actions. Moreover, it is not clear how well it will fare when crafting new recipes that it has not seen before, as needed in the tasks we consider. 

VPT~\citep{baker2022video} is an AI \mcraft agent that uses a Minecraft video gameplay dataset to learn how to navigate and interact with the world. 
MineClip~\citep{fan2022minedojo} is another AI \mcraft agent that focuses on learning a reward function to guide the agent's actions toward specific objectives. 
STEVE-1~\citep{lifshitz2024steve} is a more recent AI \mcraft agent that combines techniques from VPT and MineClip. 
Despite its advancements, STEVE-1 struggles with longer-horizon tasks like crafting and building without further training. It may also be unable to craft custom recipes to solve the \mcraft tasks we consider, such as the \pogotask in our case. 

DreamerV3~\citep{dreamerv3} is an advanced model-based reinforcement learning agent that efficiently learns long-horizon tasks in complex environments by leveraging latent world models. Instead of relying on extensive trial-and-error, the agent learned to predict future states, rewards, and environment dynamics based on past experiences, enabling it to ``imagine" different strategies in latent space. In the context of Minecraft, DreamerV3 was able to learn to mine diamonds (a very challenging task in Minecraft) using only 2.5 million frames, significantly fewer than traditional reinforcement learning methods.
In our case, where training lasts only a few thousand steps per task, even such an efficient algorithm would not be sufficient.

\section{Conclusions and Future Work}
In this work, we explored the benefits of learning a domain model for numeric planning in two settings: offline learning with expert observations and online learning. 
In particular, we compared using standard model-free algorithms for solving numeric planning problems in these settings with model-based algorithms that require learning the domain model. 
As a benchmark, we used two tasks in \mcraft, and provided both planning and \rl translation mechanisms to allow using off-the-shelf model-free and model-based algorithms, where possible. 

For the offline learning setting, we evaluated standard \il and Offline \rl algorithms, namely \bc, \gail, \dqn, and \qrdqn. 
The model-based alternative for this setting, which we called \pddla, uses \nsam~\citep{mordoch2023learning} to learn a numeric domain model of the environment and then a planner to find a plan with the learned domain model. 
Our results showed that \bc yields better results in the simpler \mcraft task, but was significantly outperformed by \pddla in the harder task. 

For the online learning setting, we evaluated standard \rl algorithms, namely, \ppo and \dqn. 
There are no algorithms for online learning of numeric domain models, and therefore, we proposed a novel hybrid strategy that we call \hybrid. 
In \hybrid, we run \rl and \nsam simultaneously, both learning from observed trajectories. The domain model gradually learned by \nsam is used to find plans as well as find shortcuts in the plans found by the \rl algorithm. 
The plans and trajectories found by each algorithm are used to improve the other,
yielding better \rl policies and more accurate domain models. 
Experimental results show that \hybrid outperforms standard \rl algorithms significantly, in terms of both success rate and quality of solution.

This work, and in particular the proposed hybrid strategy (\hybrid), opens several exciting directions for future work. 
One direction is to develop more informed action selection policies that are intentionally designed to improve the quality of the learned action model. 
This was done for classical planning~\cite{lamanna2021online} but not for numeric planning. 

Another direction for future work is to integrate \hybrid with the Go-Explore algorithm~\citep{ecoffet2021first} for deterministic and fully observable environments. 
Go-Explore is an \rl algorithm that stores promising states in the environment, and then guides the agent back to these states in later episodes. 
By replacing the memory component of the Go-Explore algorithm with an action model learning algorithm, such as \nsam, we may be able to find shortcuts to these promising states, thereby reducing the time required to reach them.

Similarly, we would like to integrate \hybrid with DreamerV3~\citep{hafner2023mastering}, a state-of-the-art Model-Based \rl algorithm for deterministic and fully observable environments. 
Instead of relying solely on real-world simulations, DreamerV3 learns the environment's dynamics (using neural networks), enabling it to perform tasks efficiently without extensive trial-and-error in the real environment. This mechanism is referred to as ``imagination''.
In this context, we propose to do ``imagination'' using a learned action model, e.g., learned using \nsam, rather than on a neural-network dynamics representation. This would allow learning from fewer steps, making the approach even more efficient.

\section*{Acknowledgments}
This work has been partially funded by the ISF grant \#1238/23 to Roni Stern. 
This work was also supported by the Israel Science Foundation (ISF) grant \#909/23 awarded to Shahaf Shperberg, by Israel's Ministry of Innovation, Science and Technology (MOST) grant \#1001706842, in collaboration with Israel National Road Safety Authority and Netivei Israel, awarded to Shahaf Shperberg, and by BSF grant \#2024614 awarded to Shahaf Shperberg.

\bibliographystyle{elsarticle-harv} 
\bibliography{main}

\begin{thebibliography}{72}
\expandafter\ifx\csname natexlab\endcsname\relax\def\natexlab#1{#1}\fi
\providecommand{\url}[1]{\texttt{#1}}
\providecommand{\href}[2]{#2}
\providecommand{\path}[1]{#1}
\providecommand{\DOIprefix}{doi:}
\providecommand{\ArXivprefix}{arXiv:}
\providecommand{\URLprefix}{URL: }
\providecommand{\Pubmedprefix}{pmid:}
\providecommand{\doi}[1]{\href{http://dx.doi.org/#1}{\path{#1}}}
\providecommand{\Pubmed}[1]{\href{pmid:#1}{\path{#1}}}
\providecommand{\bibinfo}[2]{#2}
\ifx\xfnm\relax \def\xfnm[#1]{\unskip,\space#1}\fi
\bibitem[{Abbeel and Ng(2004)}]{abbeel2004apprenticeship}
\bibinfo{author}{Abbeel, P.}, \bibinfo{author}{Ng, A.Y.}, \bibinfo{year}{2004}.
\newblock \bibinfo{title}{Apprenticeship learning via inverse reinforcement learning}, in: \bibinfo{booktitle}{Proceedings of the twenty-first international conference on Machine learning}, p.~\bibinfo{pages}{1}.
\bibitem[{Aeronautiques et~al.(1998)Aeronautiques, Howe, Knoblock, McDermott, Ram, Veloso, Weld, SRI, Barrett, Christianson et~al.}]{aeronautiques1998pddl}
\bibinfo{author}{Aeronautiques, C.}, \bibinfo{author}{Howe, A.}, \bibinfo{author}{Knoblock, C.}, \bibinfo{author}{McDermott, I.D.}, \bibinfo{author}{Ram, A.}, \bibinfo{author}{Veloso, M.}, \bibinfo{author}{Weld, D.}, \bibinfo{author}{SRI, D.W.}, \bibinfo{author}{Barrett, A.}, \bibinfo{author}{Christianson, D.}, et~al., \bibinfo{year}{1998}.
\newblock \bibinfo{title}{Pddl| the planning domain definition language}.
\newblock \bibinfo{journal}{Technical Report, Tech. Rep.} .
\bibitem[{Aineto et~al.(2019)Aineto, Celorrio and Onaindia}]{aineto2019learning}
\bibinfo{author}{Aineto, D.}, \bibinfo{author}{Celorrio, S.J.}, \bibinfo{author}{Onaindia, E.}, \bibinfo{year}{2019}.
\newblock \bibinfo{title}{Learning action models with minimal observability}.
\newblock \bibinfo{journal}{Artificial Intelligence} \bibinfo{volume}{275}, \bibinfo{pages}{104--137}.
\bibitem[{Aldinger and Nebel(2017)}]{aldinger2017interval}
\bibinfo{author}{Aldinger, J.}, \bibinfo{author}{Nebel, B.}, \bibinfo{year}{2017}.
\newblock \bibinfo{title}{Interval based relaxation heuristics for numeric planning with action costs}, in: \bibinfo{booktitle}{Joint German/Austrian Conference on Artificial Intelligence (K{\"u}nstliche Intelligenz)}, \bibinfo{organization}{Springer}. pp. \bibinfo{pages}{15--28}.
\bibitem[{Amir and Chang(2008)}]{amir2008learning}
\bibinfo{author}{Amir, E.}, \bibinfo{author}{Chang, A.}, \bibinfo{year}{2008}.
\newblock \bibinfo{title}{Learning partially observable deterministic action models}.
\newblock \bibinfo{journal}{Journal of Artificial Intelligence Research} \bibinfo{volume}{33}, \bibinfo{pages}{349--402}.
\bibitem[{Asai and Fukunaga(2018)}]{asai2018classical}
\bibinfo{author}{Asai, M.}, \bibinfo{author}{Fukunaga, A.}, \bibinfo{year}{2018}.
\newblock \bibinfo{title}{Classical planning in deep latent space: Bridging the subsymbolic-symbolic boundary}, in: \bibinfo{booktitle}{Proceedings of the aaai conference on artificial intelligence}.
\bibitem[{Baker et~al.(2022)Baker, Akkaya, Zhokov, Huizinga, Tang, Ecoffet, Houghton, Sampedro and Clune}]{baker2022video}
\bibinfo{author}{Baker, B.}, \bibinfo{author}{Akkaya, I.}, \bibinfo{author}{Zhokov, P.}, \bibinfo{author}{Huizinga, J.}, \bibinfo{author}{Tang, J.}, \bibinfo{author}{Ecoffet, A.}, \bibinfo{author}{Houghton, B.}, \bibinfo{author}{Sampedro, R.}, \bibinfo{author}{Clune, J.}, \bibinfo{year}{2022}.
\newblock \bibinfo{title}{Video pretraining (vpt): Learning to act by watching unlabeled online videos}.
\newblock \bibinfo{journal}{Advances in Neural Information Processing Systems} \bibinfo{volume}{35}, \bibinfo{pages}{24639--24654}.
\bibitem[{Benyamin et~al.(2024)Benyamin, Mordoch, Shperberg, Piotrowski and Stern}]{benyamin2024crafting}
\bibinfo{author}{Benyamin, Y.}, \bibinfo{author}{Mordoch, A.}, \bibinfo{author}{Shperberg, S.}, \bibinfo{author}{Piotrowski, W.}, \bibinfo{author}{Stern, R.}, \bibinfo{year}{2024}.
\newblock \bibinfo{title}{Crafting a pogo stick in minecraft with heuristic search}, in: \bibinfo{booktitle}{International Symposium on Combinatorial Search}, pp. \bibinfo{pages}{261--262}.
\bibitem[{Bonet and Geffner(2020)}]{bonet2020qualitative}
\bibinfo{author}{Bonet, B.}, \bibinfo{author}{Geffner, H.}, \bibinfo{year}{2020}.
\newblock \bibinfo{title}{Qualitative numeric planning: Reductions and complexity}.
\newblock \bibinfo{journal}{Journal of Artificial Intelligence Research} \bibinfo{volume}{69}, \bibinfo{pages}{923--961}.
\bibitem[{Brockman et~al.(2016)Brockman, Cheung, Pettersson, Schneider, Schulman, Tang and Zaremba}]{brockman2016openai}
\bibinfo{author}{Brockman, G.}, \bibinfo{author}{Cheung, V.}, \bibinfo{author}{Pettersson, L.}, \bibinfo{author}{Schneider, J.}, \bibinfo{author}{Schulman, J.}, \bibinfo{author}{Tang, J.}, \bibinfo{author}{Zaremba, W.}, \bibinfo{year}{2016}.
\newblock \bibinfo{title}{Openai gym}.
\newblock \bibinfo{journal}{arXiv preprint arXiv:1606.01540} .
\bibitem[{Chitnis et~al.(2021)Chitnis, Silver, Tenenbaum, Kaelbling and Lozano-P{\'e}rez}]{chitnis2021glib}
\bibinfo{author}{Chitnis, R.}, \bibinfo{author}{Silver, T.}, \bibinfo{author}{Tenenbaum, J.B.}, \bibinfo{author}{Kaelbling, L.P.}, \bibinfo{author}{Lozano-P{\'e}rez, T.}, \bibinfo{year}{2021}.
\newblock \bibinfo{title}{Glib: Efficient exploration for relational model-based reinforcement learning via goal-literal babbling}, in: \bibinfo{booktitle}{Proceedings of the AAAI Conference on Artificial Intelligence}, pp. \bibinfo{pages}{11782--11791}.
\bibitem[{Cresswell et~al.(2013)Cresswell, McCluskey and West}]{cresswell2013acquiring}
\bibinfo{author}{Cresswell, S.}, \bibinfo{author}{McCluskey, T.}, \bibinfo{author}{West, M.}, \bibinfo{year}{2013}.
\newblock \bibinfo{title}{Acquiring planning domain models using locm}.
\newblock \bibinfo{journal}{The Knowledge Engineering Review} \bibinfo{volume}{28}, \bibinfo{pages}{195--213}.
\bibitem[{Dabney et~al.(2018)Dabney, Rowland, Bellemare and Munos}]{DabneyRBM18}
\bibinfo{author}{Dabney, W.}, \bibinfo{author}{Rowland, M.}, \bibinfo{author}{Bellemare, M.G.}, \bibinfo{author}{Munos, R.}, \bibinfo{year}{2018}.
\newblock \bibinfo{title}{Distributional reinforcement learning with quantile regression}, in: \bibinfo{booktitle}{{AAAI}}, \bibinfo{publisher}{{AAAI} Press}. pp. \bibinfo{pages}{2892--2901}.
\bibitem[{Ecoffet et~al.(2021)Ecoffet, Huizinga, Lehman, Stanley and Clune}]{ecoffet2021first}
\bibinfo{author}{Ecoffet, A.}, \bibinfo{author}{Huizinga, J.}, \bibinfo{author}{Lehman, J.}, \bibinfo{author}{Stanley, K.O.}, \bibinfo{author}{Clune, J.}, \bibinfo{year}{2021}.
\newblock \bibinfo{title}{First return, then explore}.
\newblock \bibinfo{journal}{Nature} \bibinfo{volume}{590}, \bibinfo{pages}{580--586}.
\bibitem[{Fan et~al.(2022)Fan, Wang, Jiang, Mandlekar, Yang, Zhu, Tang, Huang, Zhu and Anandkumar}]{fan2022minedojo}
\bibinfo{author}{Fan, L.}, \bibinfo{author}{Wang, G.}, \bibinfo{author}{Jiang, Y.}, \bibinfo{author}{Mandlekar, A.}, \bibinfo{author}{Yang, Y.}, \bibinfo{author}{Zhu, H.}, \bibinfo{author}{Tang, A.}, \bibinfo{author}{Huang, D.A.}, \bibinfo{author}{Zhu, Y.}, \bibinfo{author}{Anandkumar, A.}, \bibinfo{year}{2022}.
\newblock \bibinfo{title}{Minedojo: Building open-ended embodied agents with internet-scale knowledge}.
\newblock \bibinfo{journal}{Advances in Neural Information Processing Systems} \bibinfo{volume}{35}, \bibinfo{pages}{18343--18362}.
\bibitem[{Fox and Long(2002)}]{fox2002pddl+}
\bibinfo{author}{Fox, M.}, \bibinfo{author}{Long, D.}, \bibinfo{year}{2002}.
\newblock \bibinfo{title}{Pddl+: Modeling continuous time dependent effects}, in: \bibinfo{booktitle}{Proceedings of the 3rd International NASA Workshop on Planning and Scheduling for Space}, p.~\bibinfo{pages}{34}.
\bibitem[{Fox and Long(2003)}]{fox2003pddl2}
\bibinfo{author}{Fox, M.}, \bibinfo{author}{Long, D.}, \bibinfo{year}{2003}.
\newblock \bibinfo{title}{Pddl2. 1: An extension to pddl for expressing temporal planning domains}.
\newblock \bibinfo{journal}{Journal of artificial intelligence research} \bibinfo{volume}{20}, \bibinfo{pages}{61--124}.
\bibitem[{Georgievski and Aiello(2015)}]{georgievski2015htn}
\bibinfo{author}{Georgievski, I.}, \bibinfo{author}{Aiello, M.}, \bibinfo{year}{2015}.
\newblock \bibinfo{title}{Htn planning: Overview, comparison, and beyond}.
\newblock \bibinfo{journal}{Artificial Intelligence} \bibinfo{volume}{222}, \bibinfo{pages}{124--156}.
\bibitem[{Gigante et~al.(2023)Gigante, Scala et~al.}]{gigante2023compilability}
\bibinfo{author}{Gigante, N.}, \bibinfo{author}{Scala, E.}, et~al., \bibinfo{year}{2023}.
\newblock \bibinfo{title}{On the compilability of bounded numeric planning.}, in: \bibinfo{booktitle}{IJCAI}, pp. \bibinfo{pages}{5341--5349}.
\bibitem[{Gnad et~al.(2023)Gnad, Helmert, Jonsson and Shleyfman}]{gnad2023planning}
\bibinfo{author}{Gnad, D.}, \bibinfo{author}{Helmert, M.}, \bibinfo{author}{Jonsson, P.}, \bibinfo{author}{Shleyfman, A.}, \bibinfo{year}{2023}.
\newblock \bibinfo{title}{Planning over integers: Compilations and undecidability}, in: \bibinfo{booktitle}{International Conference on Automated Planning and Scheduling (ICAPS)}, pp. \bibinfo{pages}{148--152}.
\bibitem[{Gonz{\'a}lez and P{\'e}rez(2009)}]{gonzalez2009improving}
\bibinfo{author}{Gonz{\'a}lez, A.}, \bibinfo{author}{P{\'e}rez, R.}, \bibinfo{year}{2009}.
\newblock \bibinfo{title}{Improving the genetic algorithm of slave}.
\newblock \bibinfo{journal}{Mathware \& Soft Computing} \bibinfo{volume}{16}, \bibinfo{pages}{59--70}.
\bibitem[{Goss et~al.(2023)Goss, Steininger, Narayanan, Oliven{\c{c}}a, Sun, Qiu, Amato, Voit, Voit and Kildebeck}]{goss2023polycraft}
\bibinfo{author}{Goss, S.A.}, \bibinfo{author}{Steininger, R.J.}, \bibinfo{author}{Narayanan, D.}, \bibinfo{author}{Oliven{\c{c}}a, D.V.}, \bibinfo{author}{Sun, Y.}, \bibinfo{author}{Qiu, P.}, \bibinfo{author}{Amato, J.}, \bibinfo{author}{Voit, E.O.}, \bibinfo{author}{Voit, W.E.}, \bibinfo{author}{Kildebeck, E.J.}, \bibinfo{year}{2023}.
\newblock \bibinfo{title}{Polycraft world ai lab (pal): An extensible platform for evaluating artificial intelligence agents}.
\newblock \bibinfo{journal}{arXiv preprint arXiv:2301.11891} .
\bibitem[{Guss et~al.(2019)Guss, Houghton, Topin, Wang, Codel, Veloso and Salakhutdinov}]{guss2019minerl}
\bibinfo{author}{Guss, W.H.}, \bibinfo{author}{Houghton, B.}, \bibinfo{author}{Topin, N.}, \bibinfo{author}{Wang, P.}, \bibinfo{author}{Codel, C.}, \bibinfo{author}{Veloso, M.}, \bibinfo{author}{Salakhutdinov, R.}, \bibinfo{year}{2019}.
\newblock \bibinfo{title}{Minerl: A large-scale dataset of minecraft demonstrations}.
\newblock \bibinfo{journal}{arXiv preprint arXiv:1907.13440} .
\bibitem[{Hafner et~al.(2023a)Hafner, Pasukonis, Ba and Lillicrap}]{hafner2023mastering}
\bibinfo{author}{Hafner, D.}, \bibinfo{author}{Pasukonis, J.}, \bibinfo{author}{Ba, J.}, \bibinfo{author}{Lillicrap, T.}, \bibinfo{year}{2023}a.
\newblock \bibinfo{title}{Mastering diverse domains through world models}.
\newblock \bibinfo{journal}{arXiv preprint arXiv:2301.04104} .
\bibitem[{Hafner et~al.(2023b)Hafner, Pasukonis, Ba and Lillicrap}]{dreamerv3}
\bibinfo{author}{Hafner, D.}, \bibinfo{author}{Pasukonis, J.}, \bibinfo{author}{Ba, J.}, \bibinfo{author}{Lillicrap, T.P.}, \bibinfo{year}{2023}b.
\newblock \bibinfo{title}{Mastering diverse domains through world models}.
\newblock \bibinfo{journal}{CoRR} \bibinfo{volume}{abs/2301.04104}.
\bibitem[{Helmert(2002)}]{helmert2002decidability}
\bibinfo{author}{Helmert, M.}, \bibinfo{year}{2002}.
\newblock \bibinfo{title}{Decidability and undecidability results for planning with numerical state variables.}, in: \bibinfo{booktitle}{AIPS}, pp. \bibinfo{pages}{44--53}.
\bibitem[{Ho and Ermon(2016)}]{ho2016generative}
\bibinfo{author}{Ho, J.}, \bibinfo{author}{Ermon, S.}, \bibinfo{year}{2016}.
\newblock \bibinfo{title}{Generative adversarial imitation learning}.
\newblock \bibinfo{journal}{Advances in neural information processing systems} \bibinfo{volume}{29}.
\bibitem[{Hoffmann(2003)}]{hoffmann2003metric}
\bibinfo{author}{Hoffmann, J.}, \bibinfo{year}{2003}.
\newblock \bibinfo{title}{The metric-ff planning system: Translating ``ignoring delete lists'' to numeric state variables}.
\newblock \bibinfo{journal}{Journal of Artificial Intelligence Research} \bibinfo{volume}{20}, \bibinfo{pages}{291--341}.
\bibitem[{Huang and Onta{\~{n}}{\'{o}}n(2022)}]{HuangO22}
\bibinfo{author}{Huang, S.}, \bibinfo{author}{Onta{\~{n}}{\'{o}}n, S.}, \bibinfo{year}{2022}.
\newblock \bibinfo{title}{A closer look at invalid action masking in policy gradient algorithms}, in: \bibinfo{booktitle}{{FLAIRS}}.
\bibitem[{Hussein et~al.(2017)Hussein, Gaber, Elyan and Jayne}]{hussein2017imitation}
\bibinfo{author}{Hussein, A.}, \bibinfo{author}{Gaber, M.M.}, \bibinfo{author}{Elyan, E.}, \bibinfo{author}{Jayne, C.}, \bibinfo{year}{2017}.
\newblock \bibinfo{title}{Imitation learning: A survey of learning methods}.
\newblock \bibinfo{journal}{ACM Computing Surveys (CSUR)} \bibinfo{volume}{50}, \bibinfo{pages}{1--35}.
\bibitem[{James et~al.(2022)James, Rosman and Konidaris}]{james2022autonomous}
\bibinfo{author}{James, S.}, \bibinfo{author}{Rosman, B.}, \bibinfo{author}{Konidaris, G.}, \bibinfo{year}{2022}.
\newblock \bibinfo{title}{Autonomous learning of object-centric abstractions for high-level planning}, in: \bibinfo{booktitle}{International Conference on Learning Representations (ICLR)}.
\bibitem[{Jang et~al.(2017)Jang, Gu and Poole}]{jang2017categorical}
\bibinfo{author}{Jang, E.}, \bibinfo{author}{Gu, S.}, \bibinfo{author}{Poole, B.}, \bibinfo{year}{2017}.
\newblock \bibinfo{title}{Categorical reparametrization with gumble-softmax}, in: \bibinfo{booktitle}{International Conference on Learning Representations (ICLR)}.
\bibitem[{Jin et~al.(2022)Jin, Ma, Jin, Zhuo, Chen and Yu}]{jin2022creativity}
\bibinfo{author}{Jin, M.}, \bibinfo{author}{Ma, Z.}, \bibinfo{author}{Jin, K.}, \bibinfo{author}{Zhuo, H.H.}, \bibinfo{author}{Chen, C.}, \bibinfo{author}{Yu, C.}, \bibinfo{year}{2022}.
\newblock \bibinfo{title}{Creativity of {AI}: Automatic symbolic option discovery for facilitating deep reinforcement learning}, in: \bibinfo{booktitle}{Proceedings of the AAAI Conference on Artificial Intelligence}, pp. \bibinfo{pages}{7042--7050}.
\bibitem[{Juba et~al.(2021)Juba, Le and Stern}]{juba2021safe}
\bibinfo{author}{Juba, B.}, \bibinfo{author}{Le, H.S.}, \bibinfo{author}{Stern, R.}, \bibinfo{year}{2021}.
\newblock \bibinfo{title}{Safe learning of lifted action models}, in: \bibinfo{booktitle}{International Conference on Principles of Knowledge Representation and Reasoning ({KR})}, pp. \bibinfo{pages}{379--389}.
\bibitem[{Juba and Stern(2022)}]{juba2022learning}
\bibinfo{author}{Juba, B.}, \bibinfo{author}{Stern, R.}, \bibinfo{year}{2022}.
\newblock \bibinfo{title}{Learning probably approximately complete and safe action models for stochastic worlds}, in: \bibinfo{booktitle}{Proceedings of the AAAI Conference on Artificial Intelligence}, pp. \bibinfo{pages}{9795--9804}.
\bibitem[{Karia et~al.(2023)Karia, Verma, Vipat and Srivastava}]{karia2023epistemic}
\bibinfo{author}{Karia, R.}, \bibinfo{author}{Verma, P.}, \bibinfo{author}{Vipat, G.}, \bibinfo{author}{Srivastava, S.}, \bibinfo{year}{2023}.
\newblock \bibinfo{title}{Epistemic exploration for generalizable planning and learning in non-stationary stochastic settings}, in: \bibinfo{booktitle}{NeurIPS 2023 Workshop on Generalization in Planning}.
\bibitem[{Lamanna et~al.(2021)Lamanna, Saetti, Serafini, Gerevini, Traverso et~al.}]{lamanna2021online}
\bibinfo{author}{Lamanna, L.}, \bibinfo{author}{Saetti, A.}, \bibinfo{author}{Serafini, L.}, \bibinfo{author}{Gerevini, A.}, \bibinfo{author}{Traverso, P.}, et~al., \bibinfo{year}{2021}.
\newblock \bibinfo{title}{Online learning of action models for pddl planning.}, in: \bibinfo{booktitle}{IJCAI}, pp. \bibinfo{pages}{4112--4118}.
\bibitem[{Lamanna and Serafini(2024)}]{Lamanna24}
\bibinfo{author}{Lamanna, L.}, \bibinfo{author}{Serafini, L.}, \bibinfo{year}{2024}.
\newblock \bibinfo{title}{Action model learning from noisy traces: a probabilistic approach}, in: \bibinfo{booktitle}{{ICAPS}}, \bibinfo{publisher}{{AAAI} Press}. pp. \bibinfo{pages}{342--350}.
\bibitem[{Le et~al.(2024)Le, Juba and Stern}]{le2024learning}
\bibinfo{author}{Le, H.S.}, \bibinfo{author}{Juba, B.}, \bibinfo{author}{Stern, R.}, \bibinfo{year}{2024}.
\newblock \bibinfo{title}{Learning safe action models with partial observability}, in: \bibinfo{booktitle}{Proceedings of the AAAI Conference on Artificial Intelligence}, pp. \bibinfo{pages}{20159--20167}.
\bibitem[{Lifshitz et~al.(2024)Lifshitz, Paster, Chan, Ba and McIlraith}]{lifshitz2024steve}
\bibinfo{author}{Lifshitz, S.}, \bibinfo{author}{Paster, K.}, \bibinfo{author}{Chan, H.}, \bibinfo{author}{Ba, J.}, \bibinfo{author}{McIlraith, S.}, \bibinfo{year}{2024}.
\newblock \bibinfo{title}{Steve-1: A generative model for text-to-behavior in minecraft}.
\newblock \bibinfo{journal}{Advances in Neural Information Processing Systems} \bibinfo{volume}{36}.
\bibitem[{Mnih et~al.(2013)Mnih, Kavukcuoglu, Silver, Graves, Antonoglou, Wierstra and Riedmiller}]{mnih2013playing}
\bibinfo{author}{Mnih, V.}, \bibinfo{author}{Kavukcuoglu, K.}, \bibinfo{author}{Silver, D.}, \bibinfo{author}{Graves, A.}, \bibinfo{author}{Antonoglou, I.}, \bibinfo{author}{Wierstra, D.}, \bibinfo{author}{Riedmiller, M.}, \bibinfo{year}{2013}.
\newblock \bibinfo{title}{Playing atari with deep reinforcement learning}.
\newblock \bibinfo{journal}{arXiv preprint arXiv:1312.5602} .
\bibitem[{Mnih et~al.(2015)Mnih, Kavukcuoglu, Silver, Rusu, Veness, Bellemare, Graves, Riedmiller, Fidjeland, Ostrovski et~al.}]{mnih2015human}
\bibinfo{author}{Mnih, V.}, \bibinfo{author}{Kavukcuoglu, K.}, \bibinfo{author}{Silver, D.}, \bibinfo{author}{Rusu, A.A.}, \bibinfo{author}{Veness, J.}, \bibinfo{author}{Bellemare, M.G.}, \bibinfo{author}{Graves, A.}, \bibinfo{author}{Riedmiller, M.}, \bibinfo{author}{Fidjeland, A.K.}, \bibinfo{author}{Ostrovski, G.}, et~al., \bibinfo{year}{2015}.
\newblock \bibinfo{title}{Human-level control through deep reinforcement learning}.
\newblock \bibinfo{journal}{nature} \bibinfo{volume}{518}, \bibinfo{pages}{529--533}.
\bibitem[{Mordoch et~al.(2023)Mordoch, Juba and Stern}]{mordoch2023learning}
\bibinfo{author}{Mordoch, A.}, \bibinfo{author}{Juba, B.}, \bibinfo{author}{Stern, R.}, \bibinfo{year}{2023}.
\newblock \bibinfo{title}{Learning safe numeric action models}, in: \bibinfo{booktitle}{{AAAI}}, \bibinfo{publisher}{{AAAI} Press}. pp. \bibinfo{pages}{12079--12086}.
\bibitem[{Mordoch et~al.(2024)Mordoch, Scala, Stern and Juba}]{MordochSSJ24}
\bibinfo{author}{Mordoch, A.}, \bibinfo{author}{Scala, E.}, \bibinfo{author}{Stern, R.}, \bibinfo{author}{Juba, B.}, \bibinfo{year}{2024}.
\newblock \bibinfo{title}{Safe learning of {PDDL} domains with conditional effects}, in: \bibinfo{booktitle}{{ICAPS}}, \bibinfo{publisher}{{AAAI} Press}. pp. \bibinfo{pages}{387--395}.
\bibitem[{Munikoti et~al.(2023)Munikoti, Agarwal, Das, Halappanavar and Natarajan}]{munikoti2023challenges}
\bibinfo{author}{Munikoti, S.}, \bibinfo{author}{Agarwal, D.}, \bibinfo{author}{Das, L.}, \bibinfo{author}{Halappanavar, M.}, \bibinfo{author}{Natarajan, B.}, \bibinfo{year}{2023}.
\newblock \bibinfo{title}{Challenges and opportunities in deep reinforcement learning with graph neural networks: A comprehensive review of algorithms and applications}.
\newblock \bibinfo{journal}{IEEE Transactions on Neural Networks and Learning Systems} .
\bibitem[{Ng and Petrick(2019)}]{ng2019incremental}
\bibinfo{author}{Ng, J.H.A.}, \bibinfo{author}{Petrick, R.P.}, \bibinfo{year}{2019}.
\newblock \bibinfo{title}{Incremental learning of planning actions in model-based reinforcement learning}, in: \bibinfo{booktitle}{IJCAI}, pp. \bibinfo{pages}{3195--3201}.
\bibitem[{Palucka(2017)}]{palucka_2017}
\bibinfo{author}{Palucka, T.}, \bibinfo{year}{2017}.
\newblock \bibinfo{title}{Polycraft world teaches science through an endlessly expansive universe of virtual gaming: https://polycraft.utdallas.edu}.
\newblock \bibinfo{journal}{MRS Bulletin} \bibinfo{volume}{42}, \bibinfo{pages}{15–17}.
\newblock \DOIprefix\doi{10.1557/mrs.2016.316}.
\bibitem[{Piotrowski and Perez(2024)}]{piotrowski2024real}
\bibinfo{author}{Piotrowski, W.}, \bibinfo{author}{Perez, A.}, \bibinfo{year}{2024}.
\newblock \bibinfo{title}{Real-world planning with pddl+ and beyond}.
\newblock \bibinfo{journal}{arXiv preprint arXiv:2402.11901} .
\bibitem[{Pomerleau(1988)}]{pomerleau1988alvinn}
\bibinfo{author}{Pomerleau, D.A.}, \bibinfo{year}{1988}.
\newblock \bibinfo{title}{Alvinn: An autonomous land vehicle in a neural network}.
\newblock \bibinfo{journal}{Advances in neural information processing systems} \bibinfo{volume}{1}.
\bibitem[{PrismarineJS(2013)}]{PrismarineJS2013}
\bibinfo{author}{PrismarineJS}, \bibinfo{year}{2013}.
\newblock \bibinfo{title}{Prismarinejs/mineflayer: Create minecraft bots with a powerful, stable, and high level javascript api.}
\bibitem[{Roberts et~al.(2017)Roberts, Piotrowski, Bevan, Aha, Fox, Long and Magazzeni}]{roberts2017automated}
\bibinfo{author}{Roberts, M.}, \bibinfo{author}{Piotrowski, W.}, \bibinfo{author}{Bevan, P.}, \bibinfo{author}{Aha, D.}, \bibinfo{author}{Fox, M.}, \bibinfo{author}{Long, D.}, \bibinfo{author}{Magazzeni, D.}, \bibinfo{year}{2017}.
\newblock \bibinfo{title}{Automated planning with goal reasoning in minecraft}, in: \bibinfo{booktitle}{ICAPS workshop on Integrated Execution of Planning and Acting (IntEx)}.
\bibitem[{Scala et~al.(2016)Scala, Haslum, Thi{\'e}baux and Ramirez}]{scala2016interval}
\bibinfo{author}{Scala, E.}, \bibinfo{author}{Haslum, P.}, \bibinfo{author}{Thi{\'e}baux, S.}, \bibinfo{author}{Ramirez, M.}, \bibinfo{year}{2016}.
\newblock \bibinfo{title}{Interval-based relaxation for general numeric planning}, in: \bibinfo{booktitle}{European Conference on Artificial Intelligence (ECAI)}, pp. \bibinfo{pages}{655--663}.
\bibitem[{Schaal(1999)}]{schaal1999imitation}
\bibinfo{author}{Schaal, S.}, \bibinfo{year}{1999}.
\newblock \bibinfo{title}{Is imitation learning the route to humanoid robots?}
\newblock \bibinfo{journal}{Trends in cognitive sciences} \bibinfo{volume}{3}, \bibinfo{pages}{233--242}.
\bibitem[{Schulman et~al.(2017)Schulman, Wolski, Dhariwal, Radford and Klimov}]{schulman2017proximal}
\bibinfo{author}{Schulman, J.}, \bibinfo{author}{Wolski, F.}, \bibinfo{author}{Dhariwal, P.}, \bibinfo{author}{Radford, A.}, \bibinfo{author}{Klimov, O.}, \bibinfo{year}{2017}.
\newblock \bibinfo{title}{Proximal policy optimization algorithms}.
\newblock \bibinfo{journal}{arXiv preprint arXiv:1707.06347} .
\bibitem[{Segura-Muros et~al.(2021)Segura-Muros, P{\'e}rez and Fern{\'a}ndez-Olivares}]{segura2021discovering}
\bibinfo{author}{Segura-Muros, J.{\'A}.}, \bibinfo{author}{P{\'e}rez, R.}, \bibinfo{author}{Fern{\'a}ndez-Olivares, J.}, \bibinfo{year}{2021}.
\newblock \bibinfo{title}{Discovering relational and numerical expressions from plan traces for learning action models}.
\newblock \bibinfo{journal}{Applied Intelligence} , \bibinfo{pages}{1--17}.
\bibitem[{Shleyfman et~al.(2023)Shleyfman, Gnad and Jonsson}]{shleyfman2023structurally}
\bibinfo{author}{Shleyfman, A.}, \bibinfo{author}{Gnad, D.}, \bibinfo{author}{Jonsson, P.}, \bibinfo{year}{2023}.
\newblock \bibinfo{title}{Structurally restricted fragments of numeric planning--a complexity analysis}, in: \bibinfo{booktitle}{AAAI Conference on Artificial Intelligence}, pp. \bibinfo{pages}{12112--12119}.
\bibitem[{Sreedharan and Katz(2023)}]{sreedharan2023optimistic}
\bibinfo{author}{Sreedharan, S.}, \bibinfo{author}{Katz, M.}, \bibinfo{year}{2023}.
\newblock \bibinfo{title}{Optimistic exploration in reinforcement learning using symbolic model estimates}.
\newblock \bibinfo{journal}{Advances in Neural Information Processing Systems} \bibinfo{volume}{36}, \bibinfo{pages}{34519--34535}.
\bibitem[{Stern and Juba(2017)}]{stern2017efficient}
\bibinfo{author}{Stern, R.}, \bibinfo{author}{Juba, B.}, \bibinfo{year}{2017}.
\newblock \bibinfo{title}{Efficient, safe, and probably approximately complete learning of action models}, in: \bibinfo{booktitle}{the International Joint Conference on Artificial Intelligence (IJCAI)}, pp. \bibinfo{pages}{4405--4411}.
\bibitem[{Sutton and Barto(2018)}]{sutton2018reinforcement}
\bibinfo{author}{Sutton, R.S.}, \bibinfo{author}{Barto, A.G.}, \bibinfo{year}{2018}.
\newblock \bibinfo{title}{Reinforcement learning: An introduction}.
\newblock \bibinfo{publisher}{MIT press}.
\bibitem[{Taitler et~al.(2024)Taitler, Alford, Espasa, Behnke, Fiser, Gimelfarb, Pommerening, Sanner, Scala, Schreiber, Segovia{-}Aguas and Seipp}]{taitler2024ipc}
\bibinfo{author}{Taitler, A.}, \bibinfo{author}{Alford, R.}, \bibinfo{author}{Espasa, J.}, \bibinfo{author}{Behnke, G.}, \bibinfo{author}{Fiser, D.}, \bibinfo{author}{Gimelfarb, M.}, \bibinfo{author}{Pommerening, F.}, \bibinfo{author}{Sanner, S.}, \bibinfo{author}{Scala, E.}, \bibinfo{author}{Schreiber, D.}, \bibinfo{author}{Segovia{-}Aguas, J.}, \bibinfo{author}{Seipp, J.}, \bibinfo{year}{2024}.
\newblock \bibinfo{title}{The 2023 international planning competition}.
\newblock \bibinfo{journal}{{AI} Mag.} \bibinfo{volume}{45}, \bibinfo{pages}{280--296}.
\bibitem[{Tang et~al.(2020)Tang, Liu, Chen and You}]{TangLCY20}
\bibinfo{author}{Tang, C.}, \bibinfo{author}{Liu, C.}, \bibinfo{author}{Chen, W.}, \bibinfo{author}{You, S.D.}, \bibinfo{year}{2020}.
\newblock \bibinfo{title}{Implementing action mask in proximal policy optimization {(PPO)} algorithm}.
\newblock \bibinfo{journal}{{ICT} Express} \bibinfo{volume}{6}, \bibinfo{pages}{200--203}.
\bibitem[{Tessler et~al.(2017)Tessler, Givony, Zahavy, Mankowitz and Mannor}]{tessler2017deep}
\bibinfo{author}{Tessler, C.}, \bibinfo{author}{Givony, S.}, \bibinfo{author}{Zahavy, T.}, \bibinfo{author}{Mankowitz, D.}, \bibinfo{author}{Mannor, S.}, \bibinfo{year}{2017}.
\newblock \bibinfo{title}{A deep hierarchical approach to lifelong learning in minecraft}, in: \bibinfo{booktitle}{Proceedings of the AAAI conference on artificial intelligence}.
\bibitem[{Verma et~al.(2023)Verma, Karia and Srivastava}]{verma2023autonomous}
\bibinfo{author}{Verma, P.}, \bibinfo{author}{Karia, R.}, \bibinfo{author}{Srivastava, S.}, \bibinfo{year}{2023}.
\newblock \bibinfo{title}{Autonomous capability assessment of sequential decision-making systems in stochastic settings}.
\newblock \bibinfo{journal}{Advances in Neural Information Processing Systems} \bibinfo{volume}{36}, \bibinfo{pages}{54727--54739}.
\bibitem[{Vinyals et~al.(2019)Vinyals, Babuschkin, Czarnecki, Mathieu, Dudzik, Chung, Choi, Powell, Ewalds, Georgiev et~al.}]{vinyals2019grandmaster}
\bibinfo{author}{Vinyals, O.}, \bibinfo{author}{Babuschkin, I.}, \bibinfo{author}{Czarnecki, W.M.}, \bibinfo{author}{Mathieu, M.}, \bibinfo{author}{Dudzik, A.}, \bibinfo{author}{Chung, J.}, \bibinfo{author}{Choi, D.H.}, \bibinfo{author}{Powell, R.}, \bibinfo{author}{Ewalds, T.}, \bibinfo{author}{Georgiev, P.}, et~al., \bibinfo{year}{2019}.
\newblock \bibinfo{title}{Grandmaster level in starcraft ii using multi-agent reinforcement learning}.
\newblock \bibinfo{journal}{Nature} \bibinfo{volume}{575}, \bibinfo{pages}{350--354}.
\bibitem[{Wang et~al.(2024)Wang, Xie, Jiang, Mandlekar, Xiao, Zhu, Fan and Anandkumar}]{wang2024voyager}
\bibinfo{author}{Wang, G.}, \bibinfo{author}{Xie, Y.}, \bibinfo{author}{Jiang, Y.}, \bibinfo{author}{Mandlekar, A.}, \bibinfo{author}{Xiao, C.}, \bibinfo{author}{Zhu, Y.}, \bibinfo{author}{Fan, L.}, \bibinfo{author}{Anandkumar, A.}, \bibinfo{year}{2024}.
\newblock \bibinfo{title}{Voyager: An open-ended embodied agent with large language models}.
\newblock \bibinfo{journal}{Trans. Mach. Learn. Res.} \bibinfo{volume}{2024}.
\bibitem[{Wang(1994)}]{wang1994learning}
\bibinfo{author}{Wang, X.}, \bibinfo{year}{1994}.
\newblock \bibinfo{title}{Learning planning operators by observation and practice}, in: \bibinfo{booktitle}{Second International Conference on Artificial Intelligence Planning Systems (AIPS)}, pp. \bibinfo{pages}{335--340}.
\bibitem[{Watkins and Dayan(1992)}]{watkins1992q}
\bibinfo{author}{Watkins, C.J.}, \bibinfo{author}{Dayan, P.}, \bibinfo{year}{1992}.
\newblock \bibinfo{title}{Q-learning}.
\newblock \bibinfo{journal}{Machine learning} \bibinfo{volume}{8}, \bibinfo{pages}{279--292}.
\bibitem[{Watkins(1989)}]{watkins1989learning}
\bibinfo{author}{Watkins, C.J.C.H.}, \bibinfo{year}{1989}.
\newblock \bibinfo{title}{Learning from delayed rewards}.
\newblock Ph.D. thesis. Oxford: King's College.
\bibitem[{Wichlacz et~al.(2019)Wichlacz, Torralba and Hoffmann}]{wichlacz2019construction}
\bibinfo{author}{Wichlacz, J.}, \bibinfo{author}{Torralba, A.}, \bibinfo{author}{Hoffmann, J.}, \bibinfo{year}{2019}.
\newblock \bibinfo{title}{Construction-planning models in minecraft}, in: \bibinfo{booktitle}{Proceedings of the ICAPS Workshop on Hierarchical Planning}, pp. \bibinfo{pages}{1--5}.
\bibitem[{Xi et~al.(2024a)Xi, Gould and Thi{\'{e}}baux}]{XiGT24}
\bibinfo{author}{Xi, K.}, \bibinfo{author}{Gould, S.}, \bibinfo{author}{Thi{\'{e}}baux, S.}, \bibinfo{year}{2024}a.
\newblock \bibinfo{title}{Neuro-symbolic learning of lifted action models from visual traces}, in: \bibinfo{booktitle}{{ICAPS}}, \bibinfo{publisher}{{AAAI} Press}. pp. \bibinfo{pages}{653--662}.
\bibitem[{Xi et~al.(2024b)Xi, Gould and Thi{\'e}baux}]{xi2024neuro}
\bibinfo{author}{Xi, K.}, \bibinfo{author}{Gould, S.}, \bibinfo{author}{Thi{\'e}baux, S.}, \bibinfo{year}{2024}b.
\newblock \bibinfo{title}{Neuro-symbolic learning of lifted action models from visual traces}, in: \bibinfo{booktitle}{Proceedings of the International Conference on Automated Planning and Scheduling}, pp. \bibinfo{pages}{653--662}.
\bibitem[{Yang et~al.(2007)Yang, Wu and Jiang}]{yang2007learning}
\bibinfo{author}{Yang, Q.}, \bibinfo{author}{Wu, K.}, \bibinfo{author}{Jiang, Y.}, \bibinfo{year}{2007}.
\newblock \bibinfo{title}{Learning action models from plan examples using weighted max-sat}.
\newblock \bibinfo{journal}{Artificial Intelligence} \bibinfo{volume}{171}, \bibinfo{pages}{107--143}.

\end{thebibliography}

\end{document}